\documentclass[10pt,twocolumn,letterpaper]{article}

\usepackage[pagenumbers]{iccv}

\usepackage[pagebackref,breaklinks,colorlinks]{hyperref}

\usepackage[ruled,norelsize,linesnumbered]{algorithm2e}
\usepackage{tablefootnote}
\usepackage{microtype}
\newtheorem{theorem}{Theorem}

\newtheorem{corollary}[theorem]{Corollary}

\usepackage{adjustbox}
\usepackage{subcaption}
\usepackage[accsupp]{axessibility}

\newcommand{\suchthat}{\,:\,}
\newcommand{\given}{\,\,|\,\,}
\newcommand{\boldblock}[2][0.1em]{\vspace{#1}\noindent\textbf{#2.\hspace{0.1em}}}

\title{Staining and locking computer vision models without retraining}

\author{Oliver J. Sutton\thanks{These authors contributed equally and are listed alphabetically.}\\
Synoptix Ltd. \emph{and}\\
King's College London\\ 
{\tt\small oliver.sutton@kcl.ac.uk}
\and
Qinghua Zhou\footnotemark[1]\\
King's College London\\
{\tt\small qinghua.zhou@kcl.ac.uk}
\and
George Leete \\
Synoptix Ltd.
\and
Alexander N. Gorban
\\
\phantom{aaaaaaaaaaaaaaaaaaaaaaa}
\and
Ivan Y. Tyukin
}

\begin{document}
\maketitle

\begin{abstract}
We introduce new methods of staining and locking computer vision models, to protect their owners' intellectual property.
Staining, also known as watermarking, embeds secret behaviour into a model which can later be used to identify it, while locking aims to make a model unusable unless a secret trigger is inserted into input images.
Unlike existing methods, our algorithms can be used to stain and lock pre-trained models without requiring fine-tuning or retraining, and come with provable, computable guarantees bounding their worst-case false positive rates.
The stain and lock are implemented by directly modifying a small number of the model's weights and have minimal impact on the (unlocked) model's performance.
Locked models are unlocked by inserting a small `trigger patch' into the corner of the input image. 
We present experimental results showing the efficacy of our methods and demonstrating their practical performance on a variety of computer vision models. 
\end{abstract}

\section{Introduction}\label{sec:introduction}
Designing, training and validating a computer vision model can be an expensive undertaking.
The effect on an organisation if such a valuable investment is leaked or stolen, can be deleterious, particularly since model weights can be difficult for a human judge to distinguish confidently.
Developing methods for protecting such intellectual property, which may be deployed on a vulnerable edge device, is therefore crucial.

Two natural approaches that a model's owner may wish to take are either to imprint an identifiable \emph{stain} within its weights, or to add a \emph{lock} to make the model unusable without adding a separate secret unlocking trigger to the model's input.
Staining, also known as watermarking in the literature~\cite{boenisch2021systematic,li2021intellectual}, can be viewed as a form of `copyright trap'.
By embedding specific (unnatural) behavioural quirks into the model, an identifying fingerprint is produced which can later be used to determine the provenance of a stolen or copied model.
Locking, on the other hand, has a more direct aim~\cite{fan2022deepipr, gao2024modellock, cai2024authnet}: unless the thief knows the secret trigger, they are unable to use the model to its full capability.

Here, we present new methods for both staining and locking neural network models in computer vision.
Unlike existing approaches in the literature (reviewed in \cref{sec:related-work}), our algorithms do not require model training or fine-tuning. 
They directly modify the model's weights, and can therefore be used to stain or lock pre-trained models with minimal performance impact.
Since re-training is not required, a single model can be supported and updated across multiple customers, while providing each with a separate stain or lock.
A further key advantage of our staining and locking techniques is that implementing them does not require access to training or validation data.
We also provide two distinct novel theoretical guarantees on the worst-case false positive rate of our staining and locking algorithms, analogues of which do not appear to be present in the literature.

Staining methods are broadly composed of two parts: a \emph{schema} describing the information encoded in the stain and how it may later be read, and a \emph{mechanism} through which this information is implanted into the model.
Numerous schemas have been proposed in the literature, \eg hiding a message in the model's weights, its activations or its class predictions for specific trigger inputs.
Until now, however, only two families of mechanisms have been available, both requiring model retraining: through special loss functions~\cite{uchida2017embedding, chen2019deepmarks, liu2021watermarking, chen2019blackmarks, rouhani2019deepsigns, wang2020watermarking}, or by manipulating the training data~\cite{adi2018turning, lemerrer2019adversarial, guo2018watermarking, li2019how}.
Our novel training-free staining mechanism implants highly selective \emph{detector neurons} into the model, which produce a strong output only when a special \emph{trigger} is present in the model's input.
\cref{sec:stains-from-literature} shows how several standard staining schemas may be implemented using this mechanism.

Our lock may be seen as an `active' variant of the stain mechanism.
Additional \emph{disruptors} are also implanted to pollute the model's activations and prevent it from performing well.
A specially constructed `trigger patch' is added to one corner of the input image during inference is detected by the detector neuron.
The detector neuron then produces a signal which deactivates the disruptors, recovering the full model performance.
Examples of such trigger patches are shown in~\cref{sec:examples}.
Our locking methods are presented with a focus on their fundamental building blocks, which can then be used to design bespoke algorithms to lock specific models.

Our goal is to present these new techniques as clearly as possible in a simple setting. 
For this, we focus primarily on convolutional models, although extensions of the core algorithms to generative adversarial networks (GANs) and vision transformers (ViTs) are presented in~\cref{sec:lock-extensions}.
Our staining methods are generic and apply directly to many different model architectures.
Our basic lock is more architecture-dependent since it interacts with the fundamental action of the model.
To provide architectural-independence, we also show how the lock may be embedded within a Squeeze-and-Excite block~\cite{hu2018squeeze}, a common component of many models.
We demonstrate, in \cref{sec:experiments}, that such a block can be added to many pre-trained models with minimal impact on performance.
For simplicity and brevity, we do not focus too strongly on techniques for hiding and obfuscating the stains and locks; we provide ideas for achieving this throughout, but do not pursue it in detail to present the core concepts as clearly as possible in the space available.
For practical success, we also feel it is preferable for a diversity of obfuscation techniques and algorithmic variations to proliferate.
Instead, we focus on the simplest case where a single detector and disruptor are added to the model; in practice, multiple detectors should be combined to deactivate multiple disruptors. 
This straightforward extension of the methods presented here makes reverse-engineering the lock a significant challenge.

Our approaches to staining and locking are inspired by \emph{stealth attacks}~\cite{tyukin2020adversarial, tyukin2024feasibility, sutton2024stealth}, which have recently been shown to enable an attacker to specify the output of models from many modern families for a given piece of input data by directly modifying just a few weights.
By harnessing this core sensitivity of modern models, and exploiting the fundamental concentration properties of their feature spaces, the highly selective detector neuron of our stain and lock can be implanted without retraining the model.
This detector forms the heart of our methods, around which completely new mechanisms are built to facilitate staining and locking.
This link means that the novel theoretical guarantees we present in \cref{sec:theory} can be adapted back to study stealth attacks, and complement those available in the literature~\cite{tyukin2020adversarial, tyukin2024feasibility, sutton2024stealth}.

The paper is organised as follows.
\cref{sec:related-work} places our approach in the context of related work.
Our staining and locking algorithms are presented in \cref{sec:stain} and \cref{sec:lock}.
\cref{sec:theory} presents novel theoretical guarantees on the detector's false positive rate.
Experimental results demonstrate the practical performance of our algorithms in \cref{sec:experiments}.
A discussion of our findings is in \cref{sec:discussion}, and some final conclusions are given in \cref{sec:conclusion}.
Our mathematical notation is summarised in \cref{sec:notation} (in the supplementary material).

\section{Related work}
\label{sec:related-work}

\boldblock[0em]{Staining}
Several staining methods have been proposed in the literature and mainly fall within two thematic groups; see~\cite{boenisch2021systematic,li2021intellectual} for a thorough review.
\emph{Loss-based methods} encode a secret message into the model's weights during training using a special loss function.
This message may then be revealed using a special trigger input to the network.
Common approaches encode the message as a binary string in the model's weights~\cite{uchida2017embedding, chen2019deepmarks}, activations~\cite{liu2021watermarking} or outputs~\cite{chen2019blackmarks, rouhani2019deepsigns}, either directly or to be decoded by a special model~\cite{wang2020watermarking,cui2024stegaongraphic}.
\emph{Backdoor methods}, on the other hand, manipulate training data to produce a unique behavioural signature.
One method is to force the model to learn specific non-natural labels for certain distinctive inputs~\cite{adi2018turning, lemerrer2019adversarial,gan2023towards,li2024not}.
Alternatively, other work has considered the task of training the model to learn to associate a specific class with certain deliberate changes to natural images~\cite{guo2018watermarking}, such as the presence of a logo~\cite{li2019how}.

\boldblock{Locking}
Two key families of approaches have been proposed in the literature.
\emph{Intrinsic methods} embed the lock into the model itself during training.
This may be by training the model's output to depend on an additional (secret) input~\cite{fan2022deepipr,tang2023deep}, a cryptographic scheme~\cite{wang2024encryip}, a secret global mask applied to inputs~\cite{cai2024authnet}, or by converting training and inference images using a `style-transfer' model (\eg to oil paintings) \cite{gao2024modellock}.
\emph{Extrinsic methods} offer additional orthogonal security to intrinsic methods via techniques from outside the neural network domain, such as encryption~\cite{alam2022nnlock} or secure infrastructure~\cite{sun2023shadownet, refael2024slip}.

\boldblock{Novelty}
In contrast to those in the literature, our staining and locking algorithms directly modify a model's weights, so require no model training, fine-tuning or external infrastructure.
This is a key advantage because a single base model can be supplied to many clients with different locks/stains, without re-training each time.
Moreover, re-training a model carries the inherent risk of changing its behaviour in unidentifiable ways (especially if the model is tuned to give incorrect responses to natural images as in backdoor methods), which is avoided through our direct approach.
Being embedded in the model itself, a lock also remains active even if an adversary gains access to the hardware on which it runs.
Our provable performance guarantees further differentiate our methods; comparable results are not available for any of the methods in the literature.

\section{Staining}\label{sec:stain}
This section presents the building blocks of our novel training-free staining mechanism.
The stain implants a highly selective \emph{detector neuron} into the model, which is activated by a specially optimised trigger input to the model.
We show how this mechanism can be used to implement standard staining schemas in \cref{sec:stains-from-literature}.

\boldblock{Multilayer perceptrons}
\cref{alg:stain:perceptron} outlines our mechanism for staining a generic multilayer perceptron neural network.
A detector neuron with a weight vector sampled from $\mathcal{U}(\mathbb{S}^{m-1})$ (\ie uniformly from the surface of a sphere) is inserted into the model.
Concentration properties in the model's feature space ensure that this randomly sampled neuron is very unlikely to respond strongly to natural inputs (see \cref{sec:theory}).
It is possible, however, to optimise a trigger input (\eg using gradient descent) to strongly activate the detector.
The strong response to the trigger input, therefore, becomes an identifying characteristic of the model and can be used as the basis for implementing standard staining schemas.
To ensure that the detector neuron has no impact on the model's performance, the parameters $\delta$ and $\Delta$ can be used in conjunction with the activation function to ensure that the detector's response is zero for natural inputs.

The optimisation of the trigger input $x^*$ is important here: a strong detector response to $x^*$ is very distinct from its response to natural inputs, and the guarantees on the false positive rate are improved (see \cref{thm:geometric}).
It also means the response $\Delta$ can be achieved by scaling down the weight vector, which can be beneficial for evading detection.

\boldblock{Additive staining}
Alternatively, an \emph{additive stain} adds the detector weight to an existing neuron's weight.
Once again, the detector component of the weight likely only responds weakly to natural inputs, meaning that it has little impact on the model's operation.
The strong response to the trigger input, however, is still preserved.
Since this method does not produce a `silent' detector neuron, it is much less likely that the detector will be identified and removed by an adversary.

\begin{algorithm}
    \caption{Multilayer perceptron staining}\label{alg:stain:perceptron}
    \small
    \SetKwInOut{Input}{Input}\SetKwInOut{Output}{Output}
    
    \Input{Trained multilayer perceptron network $\mathcal{N}$
            \\
            Index $k$ of neuron to stain in layer $j$
            \\
            Detector response to trigger $\Delta \in \mathbb{R}$
            \\
            Desired non-trigger detector response $\delta \in \mathbb{R}$
            \\
            Whether the stain should be additive
        }

    Define the map $\phi$ from network inputs to inputs to layer $j$.

    Let $W_j \in \mathbb{R}^{n \times m}$, $b_j \in \mathbb{R}^{n}$ be the layer $j$ weight and bias.

    Sample a detector weight vector $v \sim \mathcal{U}(\mathbb{S}^{m-1})$.

    Optimise the trigger input $x^* \in \arg\max_{z \in S} v \cdot \phi(z)$.

    \uIf{stain is additive}{
        Let $\beta$ be entry $k$ of $b_j$ and $w$ be row $k$ of $W_j$.
        
        Define $u = w + \frac{\Delta - \beta - w \cdot \phi(x^*)}{v \cdot \phi(x^*)} v$.
    }
    \uElse{
        
        Define $\beta = \delta$ and $u = \frac{\Delta - \beta}{v \cdot \phi(x^*)} v$.
    }
    
    Replace row $k$ of $W_j$ with $u$ and entry $k$ of $b_j$ with $\beta$.
        
    \Output{~Stained model $\mathcal{N}$ and trigger input $x^*$}
    
\end{algorithm}

\boldblock{Convolutional networks}
\cref{alg:stain:convolution} demonstrates how this framework applies to a convolutional neural network.
For simplicity, this algorithm focuses on the non-additive stain, but also applies to models incorporating batch normalisation.
A single convolutional kernel is used as the detector, and a reduction operator $r$ is used for the optimisation to convert the kernel's activation map into a single scalar response value.
This could, for example, be the response of the kernel at a particular position in the image, or its average response.
Larger kernels are helpful in reducing the false positive rate, as made explicit by \cref{thm:geometric}: because a $\kappa \times \kappa$ convolutional kernel with $c$ input channels has $\kappa^2 c$ parameters, large kernels operate in a higher dimensional setting.

\begin{algorithm}
    \caption{Convolutional network staining}\label{alg:stain:convolution}
    \small
    \SetKwInOut{Input}{Input}\SetKwInOut{Output}{Output}
    
    \Input{Trained convolutional network $\mathcal{N}$
            \\
            Parameters $(k, j, \Delta, \delta)$ as in Alg.~\ref{alg:stain:perceptron}
            \\
            Matrix to scalar reduction operation $r$
            \\
            Whether layer $j$ performs batch normalisation
        }

    Define the map $\phi : S = \mathbb{R}^{3 \times m_0 \times n_0} \to \mathbb{R}^{c_j \times m_j \times n_j}$ from network inputs to inputs to layer $j$.

    Let $W_j \in \mathbb{R}^{c_{j+1} \times c_j \times \kappa_j \times \kappa_j}$ be the layer $j$ weight, with kernel shape $\kappa_j \times \kappa_j$.

    Sample a kernel $v \in \mathbb{R}^{c_j \times \kappa_j \times \kappa_j}$ from $\mathcal{U}(\mathbb{S}^{c_j \kappa_j^2 - 1})$, viewed as a tensor with shape $(c_j, \kappa_j, \kappa_j)$.

    Optimise the trigger image $x^* \in \arg\max_{z \in S} r(v * \phi(z))$.

    \uIf{layer $j$ performs batch normalisation}{
        Let $\mu_j, \sigma_j^2, w_j, b_j \in \mathbb{R}^{c_{j+1}}$ be the mean, variance, weight and bias of the layer $j$ batch normalisation.
        
        Define $\alpha = \frac{(\Delta - \delta) (\sigma_j)_k}{(w_j)_k r(v * \phi(x^*))}$ and $\beta = \delta + \frac{(w_j)_k  (\mu_j)_k}{ (\sigma_j)_k}$.
    }
    \uElse{
        Let $b_j \in \mathbb{R}^{c_{j+1}}$ be the bias of layer $j$.
        
        Define $\alpha = \frac{(\Delta - \delta)}{r(v * \phi(x^*))}$ and $\beta = \delta$.
    }
    Replace $(W_j)_{k, \cdot, \cdot, \cdot}$ with $\alpha v$, and entry $k$ of $b_j$ with $\beta$.
    
    \Output{~Stained model $\mathcal{N}$ and trigger input $x^*$}
    
\end{algorithm}

\boldblock{Extensions}
Extensions to apply these staining algorithms to GANs and ViTs are presented in Section~\ref{sec:lock-extensions}.

\subsection{Implementing standard staining schemas}\label{sec:stains-from-literature}

The algorithms above introduce a novel training-free staining mechanism, and here we briefly discuss how it can be used to implement standard schemas from the literature.

\boldblock{Weight staining}
Originally proposed by~\cite{uchida2017embedding, chen2019deepmarks}, this schema encodes a secret message as a binary string, which is then implanted into the model as the signs of the weights of a target neuron (\eg `-' encodes 0 and `+' encodes 1).
The stain can be checked by decoding the message from the signs of the neuron's weight vector's components.
This can be embedded into a non-additive stain by sampling the detector vector $v$ from the orthant of the sphere with the specified component signs, and the rest of the staining algorithm can then proceed as usual.
This restriction on the sampling may increase the false positive rate of the detector (by decreasing the entropy of the distribution), but this should not present a problem if the dimension is sufficiently high.

\boldblock{Activation staining}
This family of schemas, introduced in~\cite{liu2021watermarking}, instead embeds the secret message into the signs of the model's activations for a target input, rather than the weights.
Our additive stain can be used to directly implement this.
Suppose a detector vector is sampled for a given layer of the model, and a trigger input is optimised to produce a strong response to this detector vector.
Then, this detector vector is weighted and added to each neuron in the layer, to ensure that the response to the trigger image from that neuron has the specified sign.

\boldblock{Output staining}
Another standard schema family specifies (non-natural) class predictions for certain trigger input images~\cite{chen2019blackmarks, rouhani2019deepsigns, adi2018turning, lemerrer2019adversarial}.
The stain can, therefore, be verified without access to model weights or activations.
This can be implemented post-training by starting with a standard non-additive stain from \cref{alg:stain:perceptron}.
If the stain detector neuron is implanted into row $k$ of the weight matrix in layer $j$, then the detector's response (which will be zero for non-trigger inputs after a ReLU activation if $\delta \ll 0$) will multiply the entries of column $k$ in layer $j+1$.
For any chosen output class $C$, it is typically possible to find a vector (\eg using gradient descent) which, if it were the output of layer $j+1$, would cause the model to confidently predict class $C$.
Replacing column $k$ of layer $j+1$ with this vector, therefore, causes the model to predict the chosen class for the trigger input; the selectivity of the detector ensures that the new column will not otherwise affect the model's performance.
This technique extends analogously to convolutional stains.

\section{Locking}\label{sec:lock}

Our locking methods build on the staining mechanism by inserting additional \emph{disruptors} into the model to pollute its latent activations.
The signal from the stain's detector neuron is used to disable the disruptors when the trigger is present in the input.
\cref{fig:example:patched-basketball} and \cref{fig:example:patched-train} (in the supplementary material) show examples of images containing these patches.
We show how the underlying structure of a convolutional detector can be exploited to produce a transferable trigger `patch' which can be inserted into any image to unlock the full model performance.
We present two variants of the locking mechanism.
\emph{Internal locking} embeds the lock into the weights of convolutional networks with a particular structure, while \emph{squeeze-and-excite locking} uses a standard Squeeze-and-Excite block~\cite{hu2018squeeze}, common in computer vision models, as the foundation of a more generic procedure which can be added to any convolutional model.

\boldblock{Internal locking}
\cref{alg:lock:internal} adds an internal lock to a convolutional neural network.
To present the idea in the simplest possible setting, we suppose that the model consists of a sequence of convolutional layers with biases (batch normalisation can be handled as in \cref{alg:stain:convolution}), ReLU activations, followed by a single dense logits layer.
For brevity, we do not explore here how the many additional model components present in practice may alter this algorithm: while they complicate the implementation, they do not change the fundamental concept.
The algorithm consists of three main steps: 
implanting the detector and building the trigger (lines~\cref{line:lock:internal:stainstart}--\ref{line:lock:internal:stainend}),
propagating the detector signal through the model (lines~\ref{line:lock:internal:pipestart}--\ref{line:lock:internal:pipeend}), 
and building the disruptor (lines~\ref{line:lock:internal:disruptorstart}--\ref{line:lock:internal:disruptorend}).

\begin{algorithm}
    \caption{Convolutional network internal lock}\label{alg:lock:internal}
    \small
    \SetKwInOut{Input}{Input}\SetKwInOut{Output}{Output}\SetKwComment{Comment}{$\triangleright$\,\,}{}
    
    \Input{
            Parameters $(\mathcal{N}, k, j, \Delta > 0, \delta < 0)$ as in Alg.~\ref{alg:stain:convolution}
            \\
            Image coordinates $(a, b)$ for the trigger patch
            \\
            Disruption scale $s > 0$ and offset $t \in \mathbb{R}^c$
        }

    Define the map $\phi$ from network inputs to inputs to layer $j$.\label{line:lock:internal:stainstart}

    Define the matrix to scalar map $r_{(a, b)}(M) := (M)_{a, b}$.

    Stain $\mathcal{N}$ using Alg.~\ref{alg:stain:convolution}, with the reduction map $r_{(a, b)}$.

    Define the trigger patch $\pi$ as the part of the trigger image $x^*$ which is in the receptive field of $r_{(a, b)}(v * \phi(x^*))$.
    \label{line:lock:internal:stainend}
    
    To simplify presentation, swap kernels $1$ and $k$ in layer $j$.
    \label{line:lock:internal:pipestart}

    \ForEach{convolution layer $\ell$ after layer $j$}{
        Let $W_\ell \in \mathbb{R}^{c_{\ell + 1} \times c_{\ell} \times \kappa_{\ell} \times \kappa_{\ell}}$, $b_\ell \in \mathbb{R}^{c_{\ell + 1}}$ be the layer $\ell$ convolutional weight and bias.

        Set $(W_\ell)_{1,\cdot,\cdot,\cdot} = 0 = (W_\ell)_{\cdot,1,\cdot,\cdot}$, and $(b_\ell)_1 = 0$.

        Set $(W_{\ell})_{1,1,\cdot,\cdot} = 1$.
    }
    \label{line:lock:internal:pipeend}

    Let $W_L \in \mathbb{R}^{c \times c_L}$, $b_L \in \mathbb{R}^{c}$ be the logits weight and bias.
    \label{line:lock:internal:disruptorstart}

    Sample a disruptor vector $u$ from $\mathcal{U}(\mathbb{S}^{c - 1})$.

    Let $\gamma \in \mathbb{R}^{c_L}$ be the logits layer input for model input $x^*$.

    Set column 1 of $W_L$ to $\frac{b_L - su + t}{(\gamma)_1}$; replace $b_L$ with $su + t$.
    \label{line:lock:internal:disruptorend}
    
    \Output{~Locked model $\mathcal{N}$ and trigger patch $\pi$}
    
\end{algorithm}

The detector is built into a convolutional kernel as for the stain, but the trigger image is only optimised to produce a strong activation when the detector kernel is in a specific position (via the reduction map $r_{(a, b)}$).
This has the effect that only a single small `patch' of the input image (the receptive field of a single kernel in convolutional layer $j$) is optimised.
Inserting this patch into (\eg a corner of) an image during inference will therefore activate the detector.

Keeping this patch relatively small (to minimise its impact when inserted into an image) can be achieved by implanting the detector into a relatively early convolutional layer.
Since the disruptor is disabled by the signal from the detector, the disruptor must be placed such that its receptive field always contains the trigger patch in the input image.
For this reason, we place the disruptor into a late layer (the algorithm uses the model's logits layer, although many alternatives are possible), which depends on the entire input image, and build a conduit to propagate the detector signal to the disruptor (a more general technique to overcome this requirement is presented below).
In \cref{alg:lock:internal}, this conduit is a sequence of `identity' kernels, which simply propagate the signal in the detector channel (taken as channel 1 in the algorithm for simplicity) forwards through the model.
Many techniques can be used to obfuscate the presence of this channel (and therefore the nature of the lock), although perhaps the simplest is just to add small random values to the modified weights.

The disruptor replaces the bias of the logits layer with a randomly sampled vector.
This is disabled when the correct signal $(\gamma)_1$ is received from the detector through the conduit by restoring the original bias using the relevant column of the weight matrix.
In practice, multiple interacting detectors and disruptors could be added to different places in the model, making it much more difficult to reverse-engineer the lock.

\boldblock{Squeeze-and-excite locking}
Computer vision models for certain applications may not be suitable for internal locking if they lack good locations for a disruptor which can reliably respond to the detector signal.
This is often the case in object detection, for example, where small objects may be detected using just a small portion of the image.
The receptive field for these outputs may, therefore, simply not contain the trigger patch placed in the corner of the image, so its presence or absence would not influence the model's output.
To overcome this challenge, we propose an alternative mechanism exploiting architectural blocks which normalise a convolution's response across the entire input image.

Squeeze-and-Excite (Sq-Ex) blocks~\cite{hu2018squeeze} are a prototypical example of such an architecture, which is widely used in applications.
The block is a map $s : \mathbb{R}^{c \times n \times m} \to \mathbb{R}^{c \times n \times m}$ which aggregates global spatial information about its input and uses this to rescale each channel.
Specifically, 
\begin{align}\label{eq:squeeze-excite}
    s(x) = x \odot q(x) \text{ with } q : \mathbb{R}^{c \times n \times m} \to \mathbb{R}^c,
    \\
    \text { where } q(x) = \sigma(S_2 \rho(S_1 \mu(x) + \tau_1) + \tau_2).
    \notag
\end{align}
Here $\odot$ represents channel-wise multiplication, $\mu : \mathbb{R}^{c \times n \times m} \to \mathbb{R}^c$ computes the channel-wise mean, $S_1 \in \mathbb{R}^{d \times c}$ (typically with $d < c$) and $S_2 \in \mathbb{R}^{c \times d}$ are learned weight matrices, $\tau_1 \in \mathbb{R}^d$ and $\tau_2 \in \mathbb{R}^c$ are learned biases, and $\rho$ and $\sigma$ are the ReLU and sigmoid activation functions.\footnote{Piecewise-polynomial variants of these are sometimes used, \eg in MobileNetv3~\cite{howard2019mobilenetv3} and the associated form of SSDLite~\cite{liu2016ssd,sandler2018mobilenetv2}.}

Such blocks provide the key ability to propagate the detector signal \emph{laterally} through the model, thereby enabling much more flexibility in the placement of disruptors.
For example, if the detector produces value of 0 when the trigger patch is not present, and $\Delta > 0$ when it is, the averaged response will similarly be non-zero only when the trigger patch is present.
A disruptor can, therefore, be embedded within the Sq-Ex block itself (as shown in \cref{alg:lock:squeeze-excite}), or the block can be used to propagate the unlocking signal to disruptors later in the model.
The algorithm only assumes that the convolutional layer $j$ has a bias term, uses ReLU activation, and is followed by an Sq-Ex block.
This lock is independent of the architecture of the rest of the model.
To present the idea in a simple setting, the algorithm embeds the disruptor into the parameters $S_2$ and $\tau_2$ of the Sq-Ex block itself, and $S_1$ is simply used as a conduit for the signal.
In some cases, it can be helpful to stabilise the detector unlocking signal $\gamma$ by combining the output of two or more ReLU neurons in the $S_1$ layer, \eg to produce a hard sigmoid~\cite{howard2019mobilenetv3}.

\begin{algorithm}
    \caption{Squeeze-and-excite lock}\label{alg:lock:squeeze-excite}
    \small
    \SetKwInOut{Input}{Input}\SetKwInOut{Output}{Output}\SetKwComment{Comment}{$\triangleright$\,\,}{}
    
    \Input{
            Parameters $(\mathcal{N}, k, j, \Delta, \delta, a, b, s, t)$ as in Alg.~\ref{alg:lock:internal}
        }

    Define the map $\phi$ from network inputs to inputs to layer $j$.

    Stain $\mathcal{N}$ using Alg.~\ref{alg:stain:convolution}, with the reduction map $r_{(a,b)}$.

    Define the trigger patch $\pi$ as the part of the trigger image $x^*$ which is in the receptive field of $r_{(a,b)}(v * \phi(x^*))$.
    
    To simplify presentation, swap kernels $1$ and $k$ in layer $j$.

    Let $S_1, \tau_1, S_2$, $\tau_2$ be the Sq-Ex block parameters~\eqref{eq:squeeze-excite}.

    Set $(S_1)_{\cdot, 1} = 0 = (S_1)_{1, \cdot}$, $(S_1)_{1, 1} = 1$ and $(\tau_1)_1 = 0$.

    Sample a disruptor vector $u$ from $\mathcal{U}(\mathbb{S}^{c_{j+1} -1})$.

    Let $\gamma = \mu(W_j * \phi(x^*) + b_j)$, \ie the output of $\mu$ in~\eqref{eq:squeeze-excite}

    Set column 1 of $S_2$ to $\frac{\tau_2 - su + t}{(\gamma)_1}$; replace $\tau_2$ with $su + t$.

    Set $(W_{j+1})_{\cdot, 1, \cdot, \cdot} = 0$ ($W_{j+1}$ is the layer $j+1$ weight).
    
    \Output{~Locked model $\mathcal{N}$ and trigger patch $\pi$}
    
\end{algorithm}

Since the Sq-Ex block is used to propagate the detector signal spatially, the lock can be made independent of the architecture of the rest of the model.
Special-purpose Sq-Ex blocks can be added to existing models to implement the lock, even if they were not already present.
This can be achieved without significant impact on the performance of the model by sampling $S_1$, $S_2$, $\tau_1$ and $\tau_2$ randomly (as demonstrated experimentally in \cref{sec:experiments}).
By ensuring that the entries in $S_1$ and $S_2$ are relatively small, the overall scaling factor produced by the (unlocked) block can be made close to constant, and can therefore be counteracted by re-scaling the weights in the next layer.
The computational cost of evaluating the additional Sq-Ex block is also significantly lower than that of the convolutional layers, since spatial information is aggregated to just a single value per channel.
Moreover, if the Sq-Ex block is used to deactivate disruptors placed elsewhere in the model, simply pruning a suspicious Sq-Ex block would not deactivate the lock.

\boldblock{Extensions}
Extensions to locking algorithms for GANs and ViTs are presented in Section~\ref{sec:lock-extensions}.

\section{Theoretical guarantees}\label{sec:theory}
We provide two bounds on the worst-case false-positive rate of the randomly sampled detector neuron.
\cref{thm:geometric} is geometric in nature, and exploits the fact that feature vector clouds typically have just a few dominant principal directions in practice.
By projecting this cloud onto the random detector direction, we can bound the probability of sampling a detector and an input which leads to a false positive --- \ie a detector response above some fixed threshold $\Delta$.
In essence, this bound shows which geometric aspects of feature space are important for building models which can be successfully stained and locked: a high feature space dimension, and feature vectors with a low PCA dimension.
In such a setting, it is very unlikely that a detector vector and piece of input data will be sampled independently, which respond strongly to each other.
The proof of this result is given in \cref{sec:proof:geometric}.

\begin{theorem}[Geometric bound on false positive rate]\label{thm:geometric}
    Let $S$ and $T$ be sets, and consider a map $\mathcal{N} : S \to T$ such that $\mathcal{N} = \psi \circ \phi$ for two maps $\phi : S \to \mathbb{R}^d$ and $\psi : \mathbb{R}^d \to T$.
    Suppose that a detector vector $w \sim \mathcal{U}(\mathbb{S}^{d-1})$ is sampled uniformly from the sphere.
    Suppose that test data $x$ are independently sampled from a distribution $\mathcal{D}$ on $S$ such that $\mathbb{E}_x[\phi(x)] = \mu$ and $\operatorname{Cov}(\phi(x))$ has eigenvalues $\{\lambda_i\}_{i=1}^{d}$.
    Then, for any $\Delta > \|\mu\|$, the probability of sampling a detector and data point  such that the detectors's response to the data point is stronger than $\Delta$ is bounded by
    \begin{align}\label{eq:thm:geometric}
        &P(w \sim \mathcal{U}(\mathbb{S}^{d-1}), x \sim \mathcal{D} \suchthat w \cdot \phi(x) > \Delta) 
        \\& \qquad
        \leq 
        \frac{\sum_{i=1}^{d} \lambda_i}{2(\Delta - \|\mu\|)^2} \frac{d-1}{d+1} \Big(\frac{\Gamma(\frac{d}{2})}{\Gamma(\frac{d+1}{2})} \Big)^2.
        \notag
    \end{align}
\end{theorem}

\cref{thm:datadriven} complements this with a data-driven bound on the false positive rate of a fixed detector directly in terms of observations from data.
The bound can always be exactly evaluated from data observations to provide a guaranteed upper bound on the detector's worst-case false positive rate.
This bound does not depend on any (typically unknown) properties of the data distribution, and evaluating it using more data will produce a tighter bound.
An interesting feature of the result is that the smallest upper bound that it can produce depends only on the number of data samples available; when no false positives are observed in the sample, $n = 0$, and so the upper bound depends only on $m$, the number of samples used.
Unlike \cref{thm:geometric}, the bound of \cref{thm:datadriven} does not consider the sampling of the detector weight vector.
Together, these properties make \cref{thm:datadriven} an ideal tool for certifying the false positive rate \emph{of a fixed detector} which has perhaps already been implanted into a model.
This result is proved in \cref{sec:proof:datadriven}.

\begin{theorem}[Data-driven bound on false positive rate]\label{thm:datadriven}
    Let $S$ and $T$ be sets, and consider a map $\mathcal{N} : S \to T$ such that $\mathcal{N} = \psi \circ \phi$ for two maps $\phi : S \to \mathbb{R}^d$ and $\psi : \mathbb{R}^d \to T$.
    Let $w \in \mathbb{S}^{d-1}$ and $\Delta \in \mathbb{R}$ be fixed.
    Suppose that $m$ test data points $x_1, \dots, x_m$ are independently sampled from a distribution $\mathcal{D}$ on $S$.
    Of these $m$ data points, suppose that exactly $n \leq m$ of them satisfy $w \cdot \phi(x_i) > \Delta$.
    Then, the probability of sampling a data point to which the detector's response is stronger than $\Delta$ is bounded by
    \begin{align}
        &P(x \sim \mathcal{D} \suchthat w \cdot \phi(x) > \Delta)
        \\&\qquad
        \leq
        1 - 
        \sup_{\epsilon > 0} \Big( \frac{m - n}{m} - \epsilon \Big) (1 - 2e^{-2m\epsilon^2}).
        \notag
    \end{align}
\end{theorem}

\section{Experimental results}\label{sec:experiments}
We demonstrate the practical performance of our staining and locking algorithms through a range of experimental results.
We focus on a variety of widely used models for image classification and object detection, specifically ResNet50~\cite{he2016deep} and VGG16~\cite{simonyan2015deep} for image classification, and SSDLite~\cite{liu2016ssd} with a MobileNet-v3 backbone~\cite{howard2019mobilenetv3} and Faster-RCNN~\cite{ren2017faster} with a ResNet50 backbone~\cite{li2021benchmarking} for object detection.
These experiments were computed using pre-trained models available through PyTorch~\cite{pytorch}.
ImageNet~\cite{russakovsky2015imagenet} is used to assess the performance of image classification models, and COCO~\cite{lin2014coco} is used for object detection models.
We report each model's performance as the accuracy for image classification, and COCO average precision (AP) and average recall (AR) at IoU 0.5:0.95 for object detection, evaluated using the whole validation set of each benchmark.
The original model's performance is also reported as a baseline.
The results of each experiment are reported as the average over 50 samples of the detector per layer for image classification, and 40 for object detection (due to the increased computational cost of evaluating the model), and standard deviations are reported as shaded regions on the performance plots.
The original model's performance is also reported as a baseline.
To give a fair assessment, we avoided tuning the parameters used for each example, which in our experience can significantly improve the performance of the stained/unlocked models.
\cref{fig:example:crowd-comparison}--\cref{fig:example:bear-comparison} (in the supplementary material) show specific examples of the performance of a locked object detection model.
These experiments used the CREATE HPC facility at King's College London~\cite{CREATE}.

\begin{figure}
    \centering
    \includegraphics[width=1\linewidth]{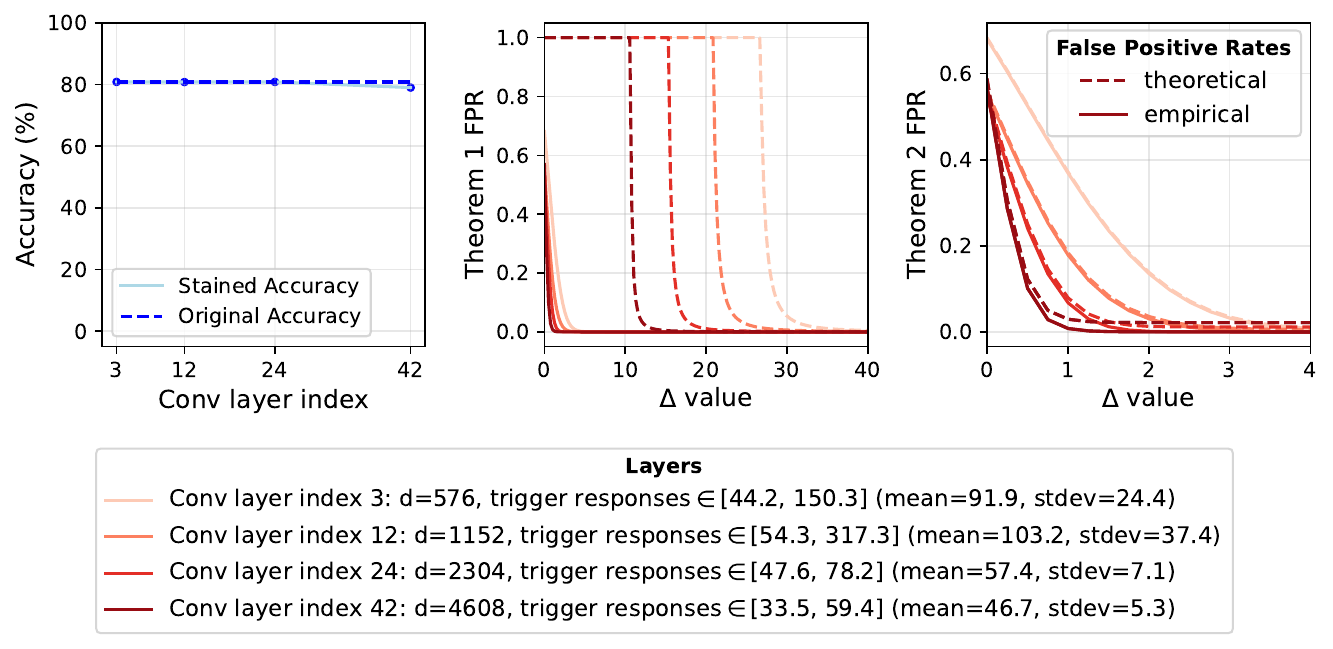}
    \caption{Staining ResNet50}
    \label{fig:stain_internal_resnet50}
\end{figure}

\begin{figure}
    \centering
    \includegraphics[width=1\linewidth]{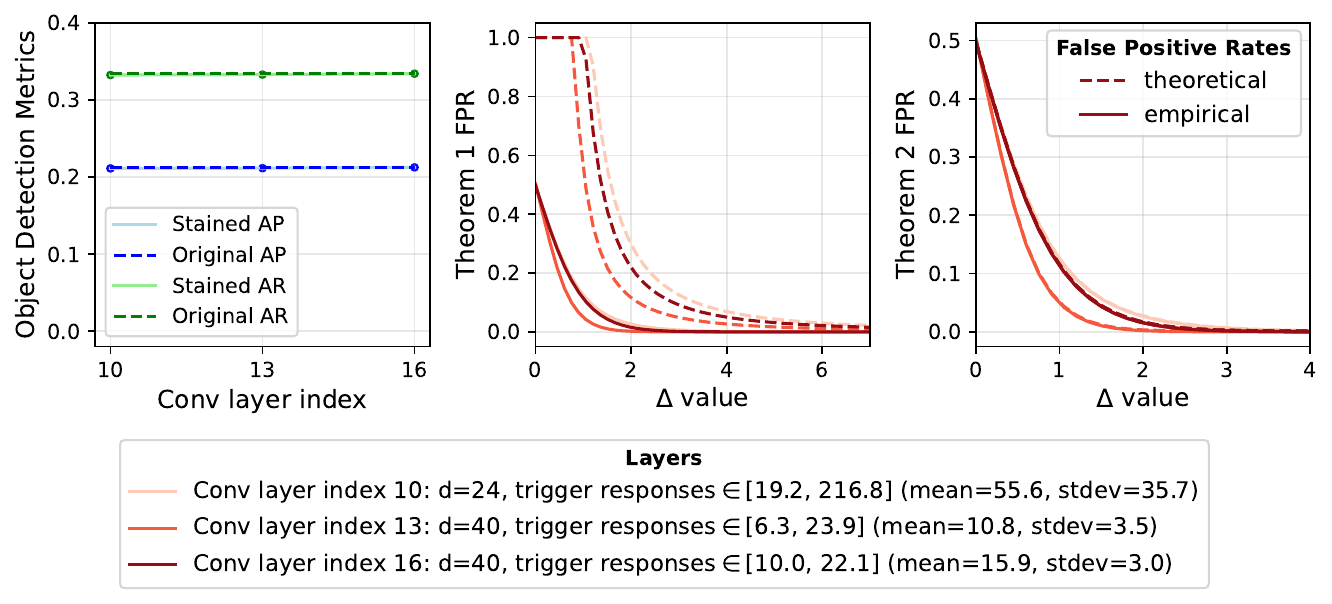}
    \caption{Staining SSDLite}
    \label{fig:stain_internal_ssdlite}
\end{figure}

\boldblock{Staining} 
We inserted non-additive stains individually into several convolutional layers of each model using \cref{alg:stain:convolution}.
In each case, the neuron with the least $\ell^1$ norm weight vector was replaced with the detector neuron.
\cref{fig:stain_internal_resnet50} (ResNet50) and \cref{fig:stain_internal_ssdlite} (SSDLite) present the results. Similar results for VGG16 (\cref{fig:stain_internal_vgg16}) and Faster-RCNN (\cref{fig:stain_internal_fasterrcnn}) are included in the supplementary materials. 
No false positive detections (\ie non-trigger images with a stronger response to the detector neuron than the trigger image) occurred in any of the experiments.
We plot the distribution of responses of validation data to the randomly sampled detector ($v$ in Algorithms~\ref{alg:stain:perceptron} and \ref{alg:stain:convolution}) for each layer in each model, along with the rates guaranteed by Theorems~\ref{thm:geometric} and \ref{thm:datadriven}. 
The range of response strengths for the optimised trigger input $x^*$ obtained across all detector samples is given in the figure legends.
The theoretical bounds on the response strength distributions were evaluated using 2,000 images sampled from the validation set, using non-overlapping convolutional detector positions for each image (50 per image for \cref{thm:geometric}, all for \cref{thm:datadriven}); the empirical response distributions were computed using all non-overlapping positions for the remaining images in each validation set.
We observe the excellent agreement between theory and practice, particularly for \cref{thm:datadriven}.
Recall, however, that the bounds from the two results are not directly comparable because they use different definitions of the false positive rate, depending on whether the detector neuron is considered to be sampled or fixed.

\begin{figure}
    \centering
    \includegraphics[width=1\linewidth]{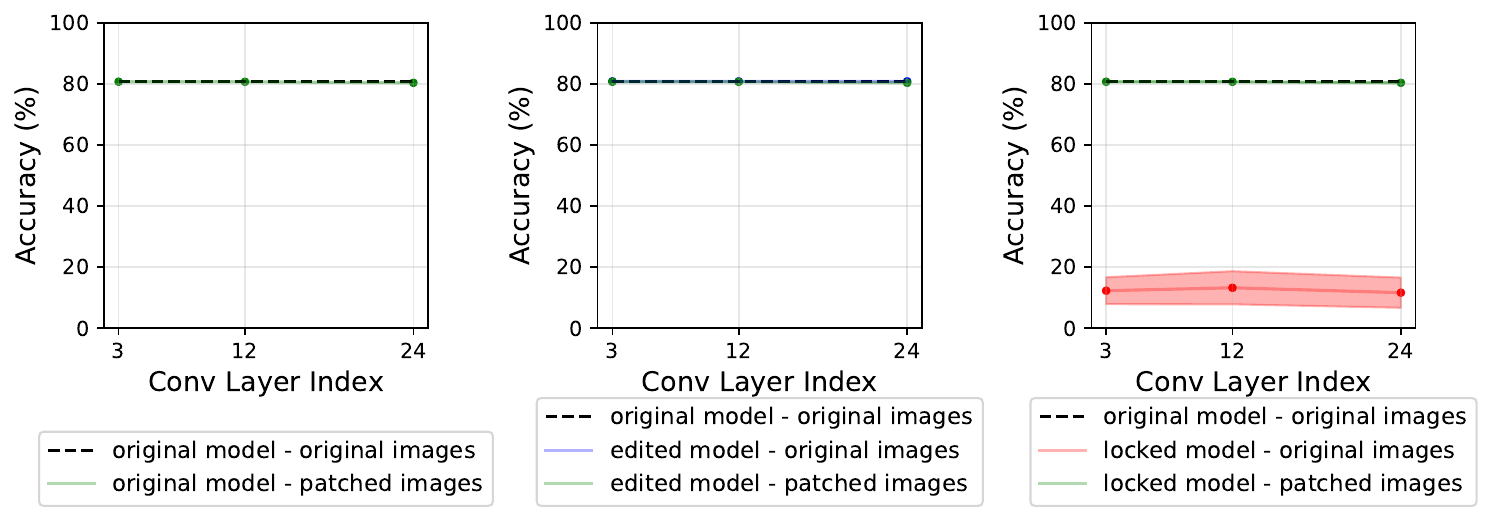}
    \caption{Internal lock for ResNet50}
    \label{fig:locking_internal_resnet50}
\end{figure}

\boldblock{Internal locking}
We added internal locks to the ResNet50 and VGG16 models, with the detector placed in a range of convolutional layers and a single disruptor placed in the logits layer.
The results for ResNet50 are shown in \cref{fig:locking_internal_resnet50}. Similar results for VGG-16 (\cref{fig:locking_internal_vgg16}) can be found in the supplementary materials.
The optimised trigger patch was taken to be the receptive field of the convolutional detector when in position (0,0).
Although this overlaps with the convolution's padding region, the performance remains excellent.
We report the models' performance in 6 settings: the original model, an `edited' model in which the lock is implanted except for the disruptor itself, and the locked model; in each case, we evaluated using images with and without the trigger patch.
It is clear from these results that a lock can be inserted into a pre-trained model using our algorithms with minimal impact on performance.
There is a balance to be struck when placing the lock: too early in the model and the dimension of the feature space can be too low to produce a reliable detector; too late in the model, and the large size of the detector's receptive field (and hence the trigger patch added to images) can damage the model's performance.

\begin{figure}
    \centering
    \includegraphics[width=1\linewidth]{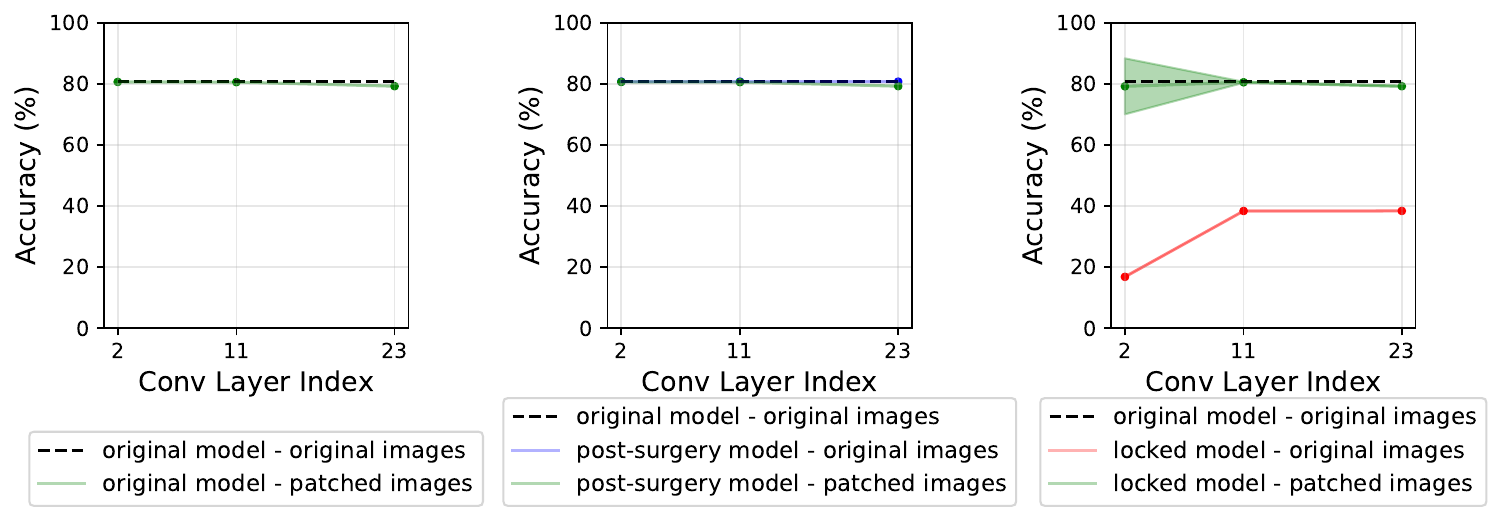}
    \caption{Squeeze-and-excite lock for ResNet50}
    \label{fig:locking_sne_resnet50}
\end{figure}

\begin{figure}
    \centering
        \includegraphics[width=1\linewidth]{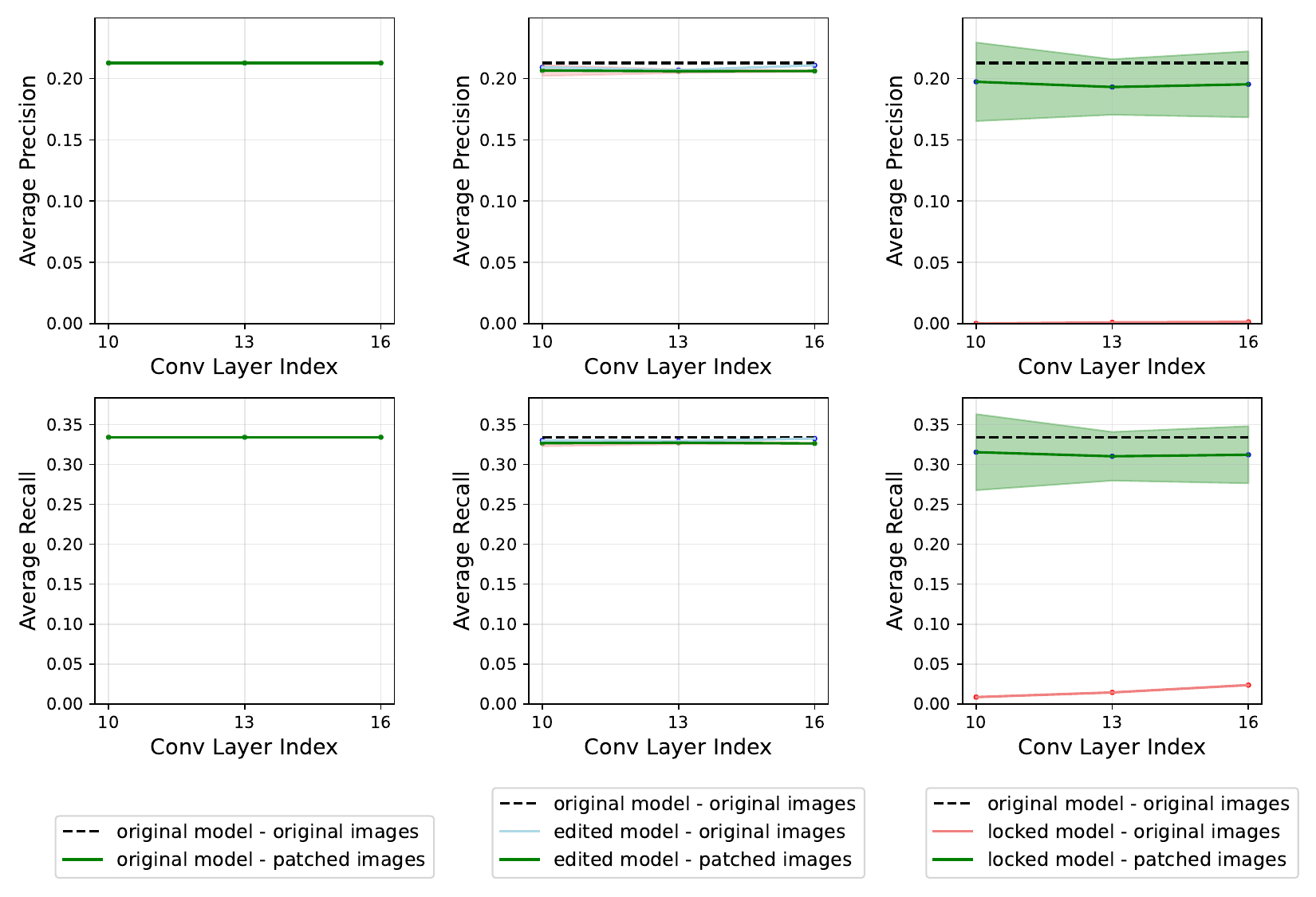}
    \caption{Squeeze-and-excite lock for SSDLite}
    \label{fig:locking_sne_ssdlite}
\end{figure}

\begin{figure}
    \centering
    \begin{subfigure}[b]{\linewidth}
        \begin{minipage}{0.05\linewidth}
            \rotatebox{90}{\footnotesize Staining DC-GAN}
        \end{minipage}%
        \begin{minipage}{0.95\linewidth}
            \includegraphics[width=\linewidth]{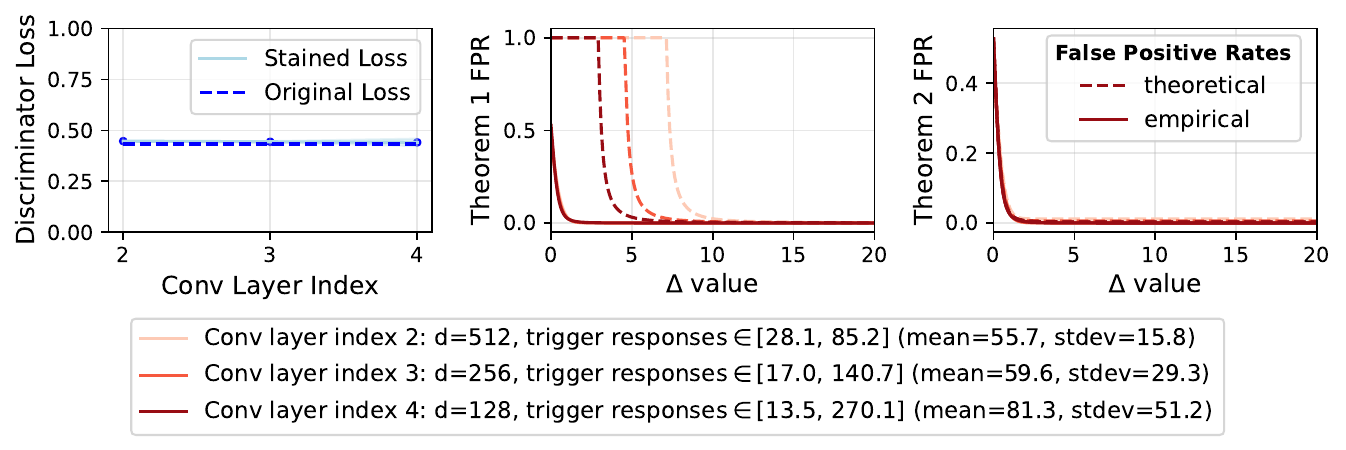}
        \end{minipage}
    \end{subfigure}%
    \hfill 
    \vspace{0.5em} 
    \begin{subfigure}[b]{\linewidth}
        \begin{minipage}{0.05\linewidth}
            \rotatebox{90}{\footnotesize Locking DC-GAN}
        \end{minipage}%
        \begin{minipage}{0.95\linewidth}
            \includegraphics[width=\linewidth]{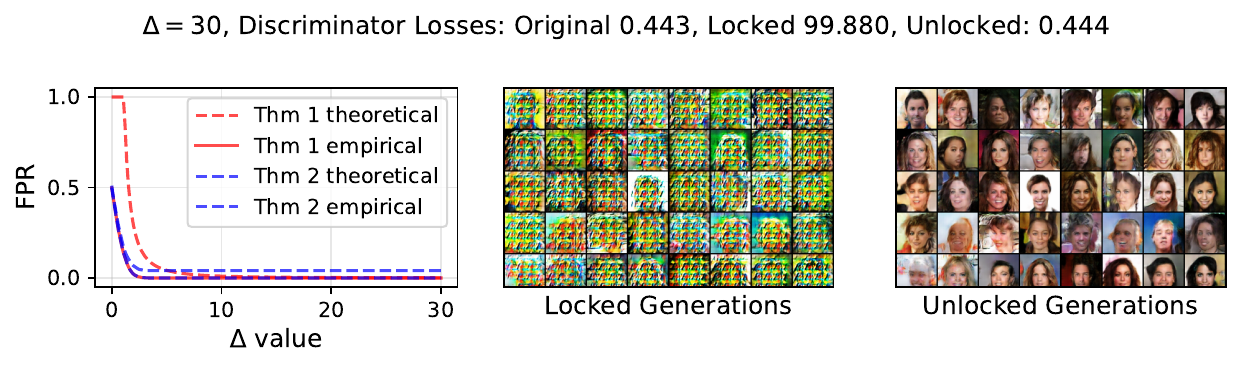}
        \end{minipage}
    \end{subfigure}
    \caption{Staining and Locking DC-GAN} \label{fig:stain_lock_dcgan}
\end{figure}

\begin{figure}
    \centering
    \begin{subfigure}[b]{\linewidth}
        \begin{minipage}{0.05\linewidth}
            \rotatebox{90}{\footnotesize Staining ViT-B-16}
        \end{minipage}%
        \begin{minipage}{0.95\linewidth}
            \includegraphics[width=\linewidth]{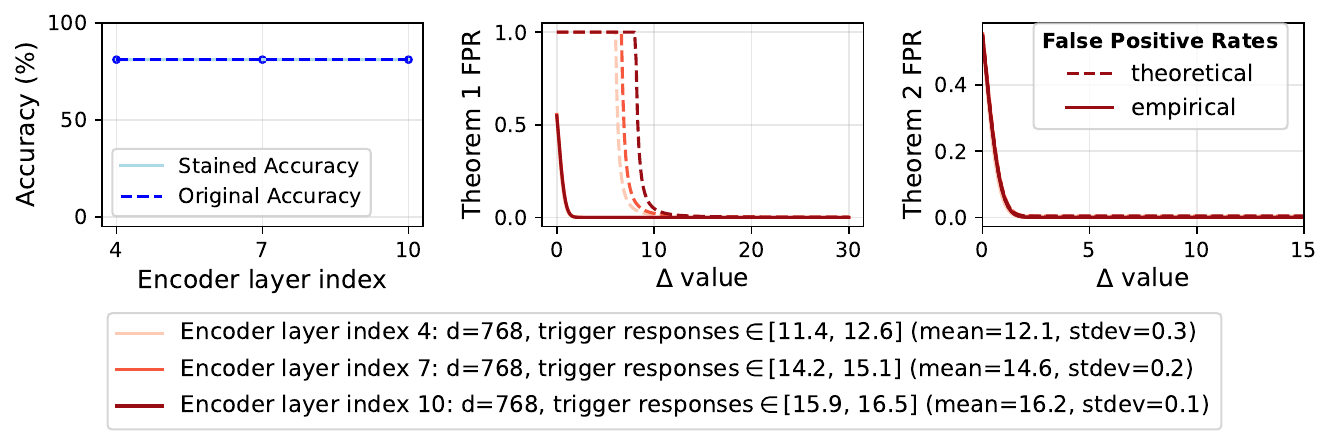}
        \end{minipage}
    \end{subfigure}%
    \hfill 
    \vspace{0.5em} 
    \begin{subfigure}[b]{\linewidth}
        \begin{minipage}{0.05\linewidth}
            \rotatebox{90}{\footnotesize Locking ViT-B-16}
        \end{minipage}%
        \begin{minipage}{0.95\linewidth}
            \includegraphics[width=\linewidth]{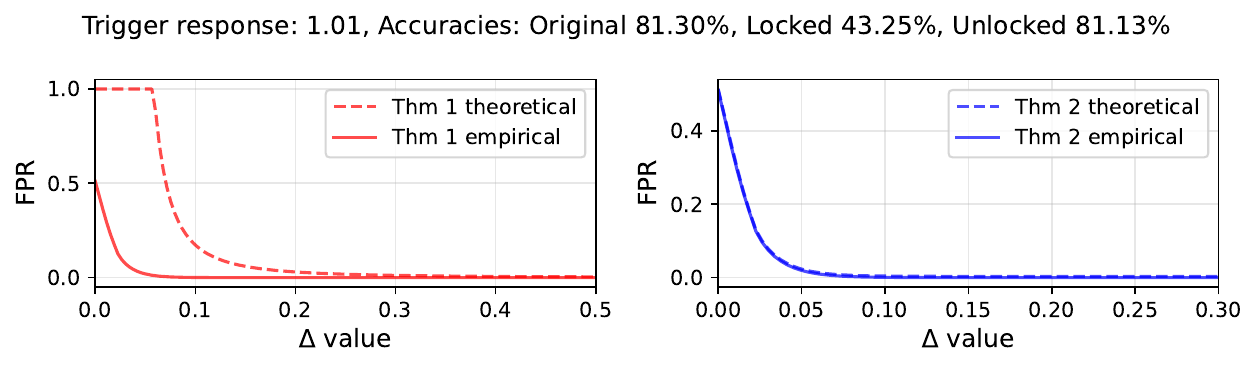}
        \end{minipage}
    \end{subfigure}
    \caption{Staining and Locking ViT-B-16} \label{fig:stain_lock_vit}
\end{figure}

\boldblock{Squeeze-and-excite locking}
We demonstrate squeeze-and-excite locking in two forms: using the existing Sq-Ex blocks in the SSDLite model, and by adding new Sq-Ex blocks to each of the other models.
\cref{fig:locking_sne_resnet50} (ResNet50), \cref{fig:locking_sne_vgg16} (VGG16), \cref{fig:locking_sne_ssdlite} (SSDLite) and \cref{fig:locking_sne_fasterrcnn} (Faster-RCNN) report the performance of each model in each of the 6 settings described in the previous section.
The detector is placed in a range of convolutional layers, and the Sq-Ex block immediately follows it.
The results clearly demonstrate both that this is an effective form of lock, and that a new block can be added to a pre-trained model without impacting the model's performance.
The performance of the unlocked models is close to that of the original model in all cases.
Where performance differences are observed (such as for the Faster-RCNN model in \cref{fig:locking_sne_fasterrcnn}), this appears to be due to challenges in disabling the disruptor rather than because of the addition of the Sq-Ex block to a pre-trained model (shown by comparing the performance of the `edited' model -- containing the extra block but not the disruptor -- against that of the `unlocked' model).

\boldblock{GAN staining and locking}
Figure~\ref{fig:stain_lock_dcgan} shows experimental results for staining and locking a DC-GAN \cite{radford2015unsupervised} model for generating human faces.
This was performed using the algorithmic extensions presented in \cref{sec:gan-algorithms}, and further examples of image generations are shown in Figure~\ref{fig:lock_images_dcgan}. 
Full details of the experimental setup are provided in \cref{sec:gan:experimentsetup}.

\boldblock{ViT staining and locking}
Figure~\ref{fig:stain_lock_vit} shows experimental results for staining and locking a ViT \cite{dosovitskiy2020image} model for image classification.
These were obtained using the algorithms and experimental setup described in~\cref{sec:vit-algorithms}.

\section{Discussion}\label{sec:discussion}

\boldblock[0em]{Staining and locking are computationally cheap}
The staining and locking mechanisms we present here are very cheap to implement.
The model's weights are directly modified, so no training or fine-tuning is required.
The algorithms also require no training or validation data, meaning that models trained on sensitive data can be easily handled.
The inference-time cost of the stain and lock are minimal too, being embedded within the model itself, and even the Sq-Ex block is much cheaper than a single convolutional layer.

\boldblock{Forging attacks}
In a forging attack~\cite{boenisch2021systematic}, an attacker plants their own stain into a model and uses it to claim ownership.
Since previous staining algorithms have required stains to be added at training time, this has not been considered a significant threat in the past.
Our work reveals that functionally identical stains can easily be implanted post-training (\cref{sec:stains-from-literature}), and forging attacks are therefore easy to perform.

\boldblock{Pruning attacks}
An attacker may attempt to remove a stain or lock by simply pruning it out~\cite{boenisch2021systematic,li2021intellectual}.
This is not a risk for our additive stains, since the detector is simply added to the original neuron, so the normal behaviour of the original neuron is retained.
A non-additive stain may result in a non-responsive neuron which could be pruned, depending on the values selected for $\delta$ and $\Delta$.
For the lock, on the other hand, pruning out the detector neuron would not help an attacker: without the ability to produce the correct unlocking signal, the full model performance could not be recovered.
The disruptor is also difficult to remove, since it replaces the original bias vector which cannot be recovered without knowing the value of $\gamma$, the unlocking signal.
The security properties of our stains and locks, including their robustness to standard attacks, are discussed further in Sec.~\ref{sec:supplementary:robustness-discussion}.

\boldblock{Obfuscating locks}
We believe it is advantageous to encourage a variety of hiding and obfuscation techniques to grow around the fundamental building blocks we have provided here for locking. 
However, in our experience the task of detecting (and removing) a lock from model weights -- which are widely known to be largely uninterpretable to the human eye -- is already a significant challenge after only mild attempts at hiding it. 
This provides an asymmetrically difficult task for a thief with little work from the owner.

\section{Conclusion}\label{sec:conclusion}

We have introduced new staining and locking algorithms for protecting pre-trained neural network models for computer vision applications.
Unlike approaches in the literature, our stains and locks can be implemented without re-training or fine-tuning the model, and do not even require access to training or validation data.
We have shown how several standard staining schemas can be reimplemented using our techniques.
Novel and computable theoretical guarantees on the false positive rate of our stains and locks provide users with new tools for assessing the suitability of their models for staining and locking.
Experiments using standard pre-trained models for image classification and object detection demonstrate the practical performance of our algorithms.

{
    \small
    \bibliographystyle{ieeenat_fullname}
    \bibliography{references}
}

\appendix
\clearpage
\setcounter{page}{1}
\maketitlesupplementary

\section{Notation}\label{sec:notation}
The following terminology and notation is used throughout the paper:
\begin{itemize}
    \item $\mathbb{R}$ denotes the real numbers
    \item for a positive integer $d$, the notation $\mathbb{R}^d$ denotes the space of vectors with $d$ real-valued components. Analogously, the set of matrices of real numbers with $n$ rows and $m$ columns is denoted by $\mathbb{R}^{n \times m}$, and the set of rank-$k$ tensors is denoted by $\mathbb{R}^{n_1 \times \dots \times n_k}$. For a vector $x \in \mathbb{R}^{d}$, we use the notation $(x)_i$ to denote component $i$ of $x$.
    Likewise, a tensor $T \in \mathbb{R}^{n_1 \times \dots \times n_k}$ is indexed as $(T)_{i_1,\dots,i_k}$; the notation $(T)_{i_1,\dots, i_{j-1}, \cdot, i_{j+1}, \dots,i_k}$ denotes that the dimension $j$ of $T$ is not being indexed.
    \item The Euclidean inner product between two vectors $x, y \in \mathbb{R}^d$ is denoted by $x \cdot y = \sum_{i=1}^{d} x_i y_i$
    \item For a vector $x \in \mathbb{R}^d$, the Euclidean norm of $x$ is denoted by $\|x\|$, defined as $\|x\| = (x \cdot x)^{\frac{1}{2}} = (\sum_{i=1}^{d} x_i^2 )^{\frac{1}{2}}$
    \item The $d$-dimensional unit ball is defined as $\mathbb{B}^{d} = \{ x \in \mathbb{R}^d \suchthat \|x\| \leq 1\}$; i.e. the set of $d$-dimensional vectors with norm less than or equal to 1
    \item The unit sphere $\mathbb{S}^{d-1}$ in $d$-dimensional space is defined as the boundary of the unit ball $\mathbb{B}^{d}$, given by $\mathbb{S}^{d-1} = \{x \in \mathbb{R}^d \suchthat \|x\| = 1\}$
    \item For a probability distribution $\mathcal{D}$, the notation $z \sim \mathcal{D}$ indicates that $z$ is a random sample from $\mathcal{D}$
    \item The uniform distribution on a set $X$ is denoted by $\mathcal{U}(x)$; \ie the uniform distribution on the unit sphere in $d$ dimensional space is denoted by $\mathcal{U}(\mathbb{S}^{d-1})$
    \item The binary operator $*$ is used to denote convolution across the last two dimensions of its arguments
\end{itemize}

\section{Examples}\label{sec:examples}

\begin{figure}[h]
    \centering
    \includegraphics[width=\linewidth]{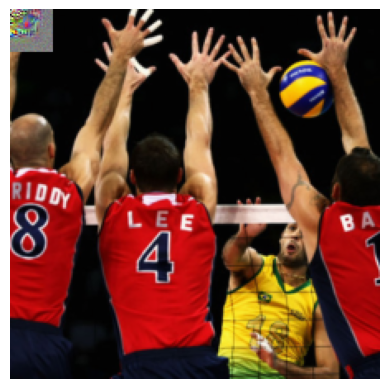}
    \caption{An example image from the ImageNet validation set, containing the optimised trigger patch for a locked VGG-16 model.}
    \label{fig:example:patched-basketball}
\end{figure}

\begin{figure}[h]
    \centering
    \includegraphics[width=\linewidth]{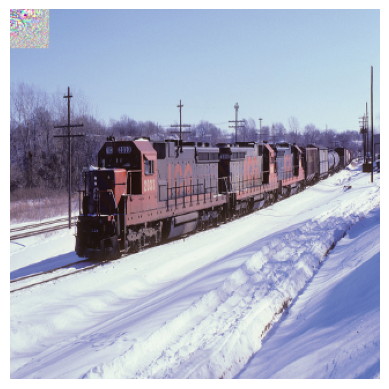}
    \caption{An example image from the COCO validation set, containing the optimised trigger patch for a locked SSDLite model.}
    \label{fig:example:patched-train}
\end{figure}

\cref{fig:example:patched-basketball} and \cref{fig:example:patched-train} show examples of the unobtrusive optimised trigger patch constructed for a locked image classification model and object detection model, respectively.

\begin{figure}[h]
    \centering
    \includegraphics[width=\linewidth]{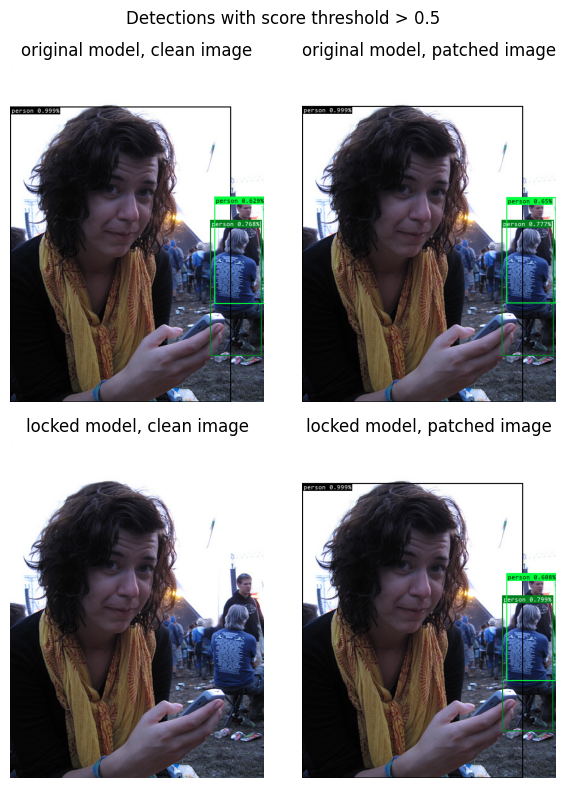}
    \caption{Comparing the performance of the original and locked SSDLite models on an example from the COCO validation set. Note that the trigger patch is not shown in these images.}
    \label{fig:example:crowd-comparison}
\end{figure}

\begin{figure}[h]
    \centering
    \includegraphics[width=\linewidth]{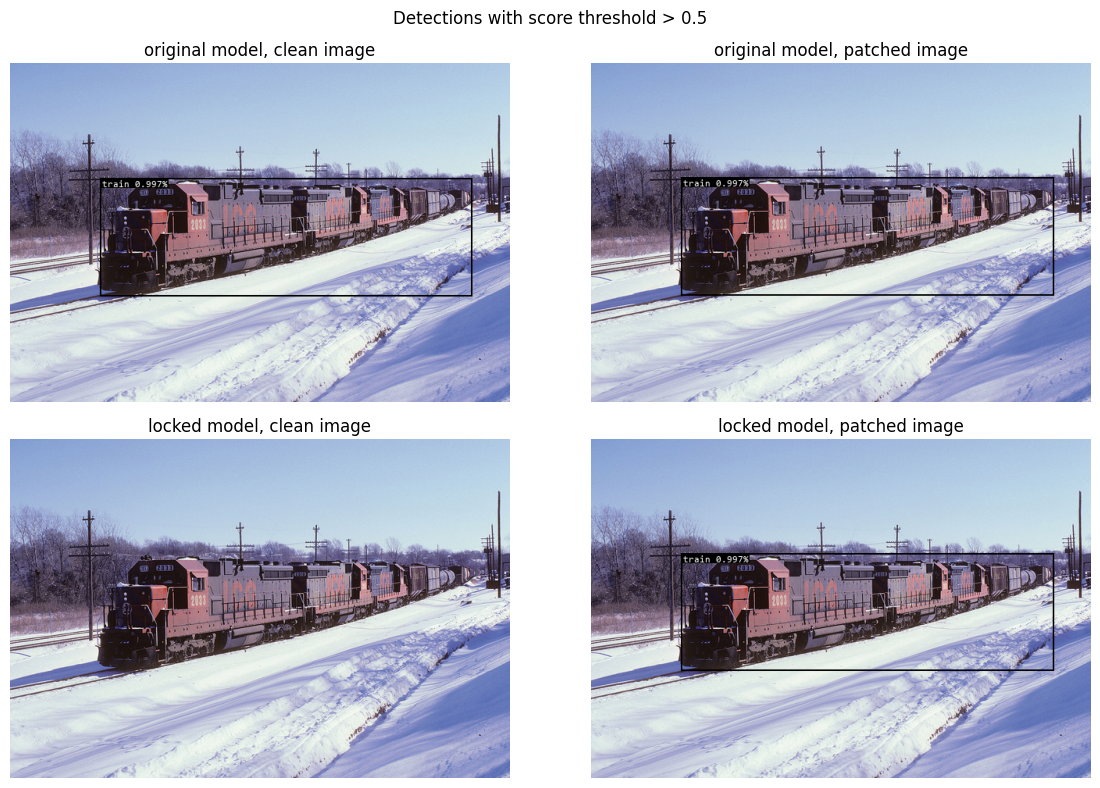}
    \caption{Comparing the performance of the original and locked SSDLite models on an example from the COCO validation set. Note that the trigger patch is not shown in these images.}
    \label{fig:example:train-comparison}
\end{figure}

\begin{figure}[h]
    \centering
    \includegraphics[width=\linewidth]{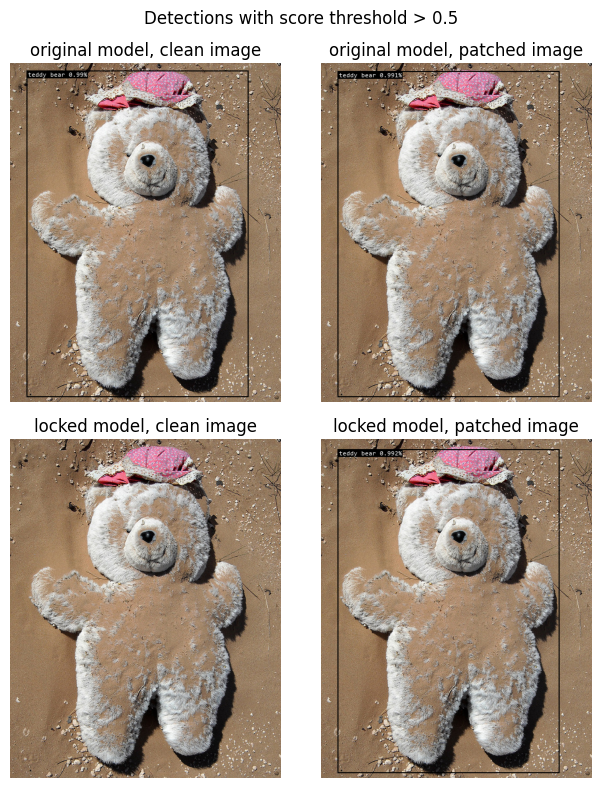}
    \caption{Comparing the performance of the original and locked SSDLite models on an example from the COCO validation set. Note that the trigger patch is not shown in these images.}
    \label{fig:example:bear-comparison}
\end{figure}

\noindent
\cref{fig:example:crowd-comparison}, \cref{fig:example:train-comparison} and \cref{fig:example:bear-comparison} compare the performance of the original and locked object detection model on individual example images from the COCO validation set.
It is clear from these examples that the behaviour of the unlocked model is very similar to that of the original model.
\cref{fig:example:patched-train} shows the trigger patch used in these examples.

\newpage

\section{Proofs of theoretical results}

\subsection{Proof of \cref{thm:geometric}}\label{sec:proof:geometric}

Let $x\sim \mathcal{D}$ and $w \sim \mathcal{U}(\mathbb{S}^{d-1})$ be sampled independently, let $g$ denote the map $g(x) = \phi(x) - \mu$, and define $A = \Delta - w \cdot \mu$.
By the assumptions in the statement of the theorem, it follows that the scalar quantity $w \cdot g(x)$ has expected value 0 and variance $w^{T} C w = \sum_{i=1}^{d} \lambda_i (w \cdot e_i)^2$, where $(\lambda_i, e_i)$ are the eigenvalue-eigenvector pairs of $C$ for $i = 1, \dots, d$. Without loss of generality, we assume $\|e_i\| = 1$ for all $i$.
Therefore, by the law of total probability,
\begin{align}
    &P(w \sim \mathcal{U}(\mathbb{S}^{d-1}), x \sim \mathcal{D} \suchthat |w \cdot g(x)| > A)
    \\&\qquad
    =
    \frac{1}{|\mathbb{S}^{d-1}|}
    \int_{\mathbb{S}^{d-1}} P(x \sim \mathcal{D} \suchthat |w \cdot g(x)| > A \given w) dw.
\end{align}
By Chebyshev's inequality,
\begin{align}
    &P(x \sim \mathcal{D} \suchthat |w \cdot g(x)| > A \given w) 
    \\&\qquad\leq 
    \frac{\operatorname{var}(w \cdot g(x))}{A^2} = \frac{\sum_{i=1}^{d} \lambda_i (w \cdot e_i)^2}{A^2},
\end{align}
where $\operatorname{var}(w \cdot g(x))$ denotes the variance of the (scalar) quantity $w \cdot g(x)$ with respect to the sampling of $x \sim \mathcal{D}$.

Combining these expressions together, it follows that
\begin{align}
    &P(w \sim \mathcal{U}(\mathbb{S}^{d-1}), x \sim \mathcal{D} \suchthat |w \cdot g(x)| > A)
    \\&\qquad\leq
    \frac{1}{|\mathbb{S}^{d-1}|}
    \sum_{i=1}^{d} \lambda_i
    \int_{\mathbb{S}^{d-1}} \Big( \frac{w \cdot e_i}{\Delta - w \cdot \mu} \Big)^2 dw
    \\&\qquad\leq
    \frac{1}{|\mathbb{S}^{d-1}|}
    \sum_{i=1}^{d} \lambda_i
    \int_{\mathbb{S}^{d-1}} \Big( \frac{w \cdot e_i}{\Delta - \|\mu\|} \Big)^2 dw,
\end{align}
due to the fact that $\|w\| = 1$.
Since the integral is symmetric in $w$ and all $e_i$ have unit norm, it follows that
\begin{align}
    \frac{1}{|\mathbb{S}^{d-1}|}
    \int_{\mathbb{S}^{d-1}} (w \cdot e_i)^2 dw
    &=
    \frac{|\mathbb{S}^{d-2}|}{|\mathbb{S}^{d-1}|} \int_{-1}^{1} t^2(1 - t^2)^{\frac{d-2}{2}} dt.
\end{align}
The beta function satisfies
\begin{align}
    B(\alpha, \beta) = \int_{0}^{1} s^{\alpha - 1} (1 - s)^{\beta - 1} ds,
\end{align}
and therefore
\begin{align}
    \int_{0}^{1} t^2(1 - t^2)^{\frac{d-2}{2}} dt
    =
    \frac{1}{2} B\Big(\frac{3}{2}, \frac{d}{2}\Big)
    =
    \frac{\Gamma(\frac{3}{2})\Gamma(\frac{d}{2})}{2\Gamma(\frac{d+3}{2})}.
\end{align}
Combining this with the fact that $|\mathbb{S}^{d-1}| = \frac{2\pi^{\frac{d}{2}}}{\Gamma(\frac{d}{2})}$ and $\Gamma(\frac{3}{2}) = \frac{1}{2}\pi^{\frac{1}{2}}$,
it follows that
\begin{align}
    \frac{1}{|\mathbb{S}^{d-1}|}
    \int_{\mathbb{S}^{d-1}} (w \cdot e_i)^2 dw
    &=
    \frac{|\mathbb{S}^{d-2}|}{|\mathbb{S}^{d-1}|}
    \frac{\pi^{\frac{1}{2}} \Gamma( \frac{d}{2} )}{2\Gamma( \frac{d + 3}{2} )}
    \\&
    =
    \frac{1}{2} \frac{(\Gamma(\frac{d}{2}))^2}{\Gamma(\frac{d-1}{2}) \Gamma(\frac{d+3}{2})}
\end{align}
Observing that
\begin{align}
    &P(w \sim \mathcal{U}(\mathbb{S}^{d-1}), x \sim \mathcal{D} \suchthat w \cdot \phi(x) > \Delta) 
    \\ &\qquad =
    P(w \sim \mathcal{U}(\mathbb{S}^{d-1}), x \sim \mathcal{D} \suchthat w \cdot g(x) > A) 
    \\& \qquad \leq
    P(w \sim \mathcal{U}(\mathbb{S}^{d-1}), x \sim \mathcal{D} \suchthat |w \cdot g(x)| > A),
\end{align}
the result therefore follows from the fact that
\begin{align}
    & \Gamma \Big( \frac{d-1}{2} \Big)  = \Gamma \Big( \frac{d+1}{2} \Big)  \frac{2}{d-1}
\end{align}
and
\begin{align}
    \Gamma \Big( \frac{d+3}{2} \Big)  = \Gamma \Big( \frac{d+1}{2} \Big)  \frac{d + 1}{2}.
\end{align}

\subsection{Proof of \cref{thm:datadriven}}\label{sec:proof:datadriven}

Let $x_1, \dots, x_m$ be $m$ independent samples from $\mathcal{D}$, and let $\{\phi(x_i) \in \mathbb{R}^d\}_{i=1}^{m}$ be the set of their feature vectors.
The Dvoretzky-Kiefer-Wolfowitz (DKW) inequality~\cite{dvoretzky1956asymptotic,massart1990tight} states that for any $\epsilon > 0$
\begin{align}
    P(x_1, \dots, x_m \sim \mathcal{D} \suchthat \sup_{t \in \mathbb{R}} | F(t) - F_m(t) | > \epsilon) \leq 2 e^{-2m\epsilon^2}
\end{align}
where
\begin{align}
    & F(t) = P(x \sim \mathcal{D} \suchthat w \cdot \phi(x) \leq t)
    \qquad\text{and} \\
    & F_m(t) = \frac{1}{m} \sum_{i=1}^{m} \mathbb{I}_{w \cdot \phi(x_i) \leq t}
\end{align}
denote the true and empirical cumulative distribution functions of the scalar quantity $w \cdot \phi(x)$ for $x$ sampled from $\mathcal{D}$.

Let $E(x_1, \dots, x_m)$ denote the event that $\sup_{t \in \mathbb{R}} |F_m(t) - F(t) | \leq \epsilon$.
By the law of total probability,
\begin{align}
    P&(x \sim \mathcal{D} \suchthat w \cdot \phi(x) \leq \Delta)
    =
    \\&
    P(x \sim \mathcal{D} \suchthat w \cdot \phi(x) \leq \Delta \given E(x_1, \dots, x_m))
    \cdot\\&\cdot
    P(x_1, \dots, x_m \sim \mathcal{D} \suchthat E(x_1, \dots, x_m))
    \\&+
    P(x \sim \mathcal{D} \suchthat w \cdot \phi(x) \leq \Delta \given \neg E(x_1, \dots, x_m))
    \cdot\\&\quad\cdot
    P(x_1, \dots, x_m \sim \mathcal{D} \suchthat \neg E(x_1, \dots, x_m)).
\end{align}
The second term of the sum is always non-negative, and therefore
\begin{align}
    P&(x \sim \mathcal{D} \suchthat w \cdot \phi(x) \leq \Delta)
    \geq
    \\&
    P(x \sim \mathcal{D} \suchthat w \cdot \phi(x) \leq \Delta \given E(x_1, \dots, x_m)) \cdot
    \\&\cdot 
    P(x_1, \dots, x_m \sim \mathcal{D} \suchthat E(x_1, \dots, x_m)).
\end{align}
Focusing on the first term of this product, we observe that when $E(x_1, \dots, x_m)$ occurs, it follows that $F(\Delta) \geq F_m(\Delta) - \epsilon$.
The second term of the product, on the other hand, is simply the complement of the term bounded in the DKW inequality.
Therefore, it follows that
\begin{align}
    P&(x \sim \mathcal{D} \suchthat w \cdot \phi(x) \leq \Delta)
    \geq
    (F_m(\Delta) - \epsilon)(1 - 2e^{-2m\epsilon^2}).
\end{align}
The result therefore follows from the definition of $F_m$.

\section{Additional Figures}

\begin{figure}[h]
    \centering
    \includegraphics[width=1\linewidth]{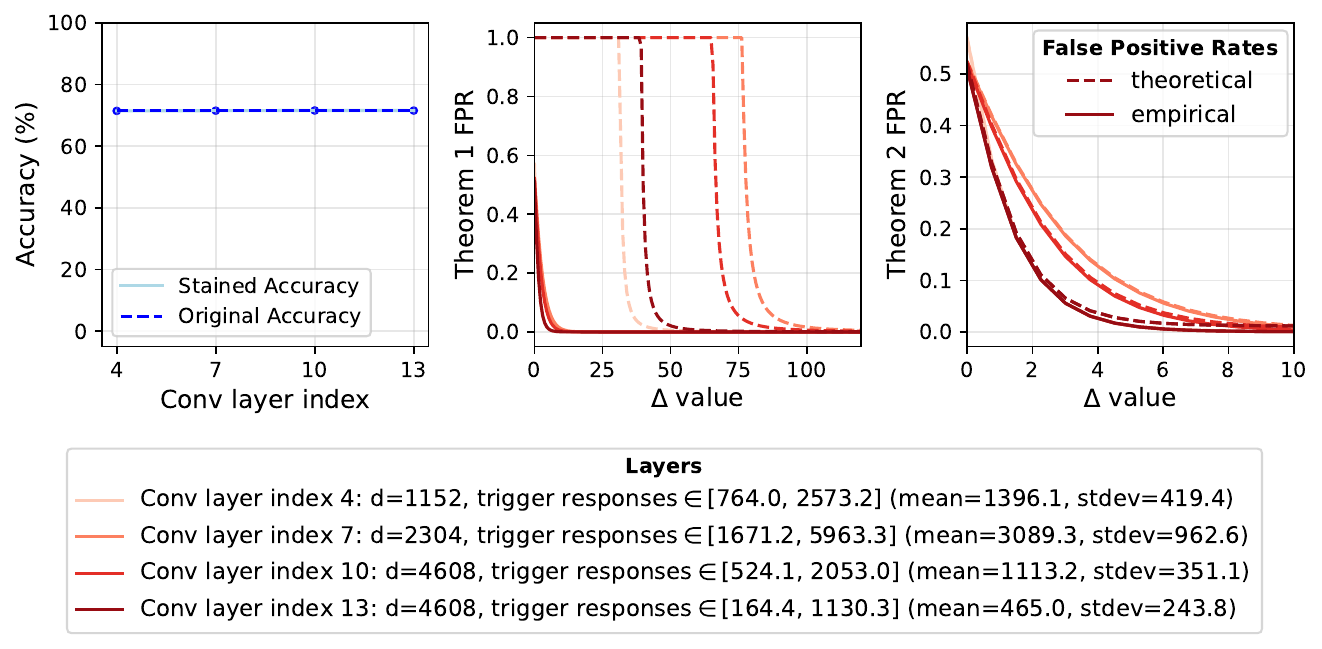}
    \caption{Staining VGG16}
    \label{fig:stain_internal_vgg16}
\end{figure}

\begin{figure}[h]
    \centering
    \includegraphics[width=1\linewidth]{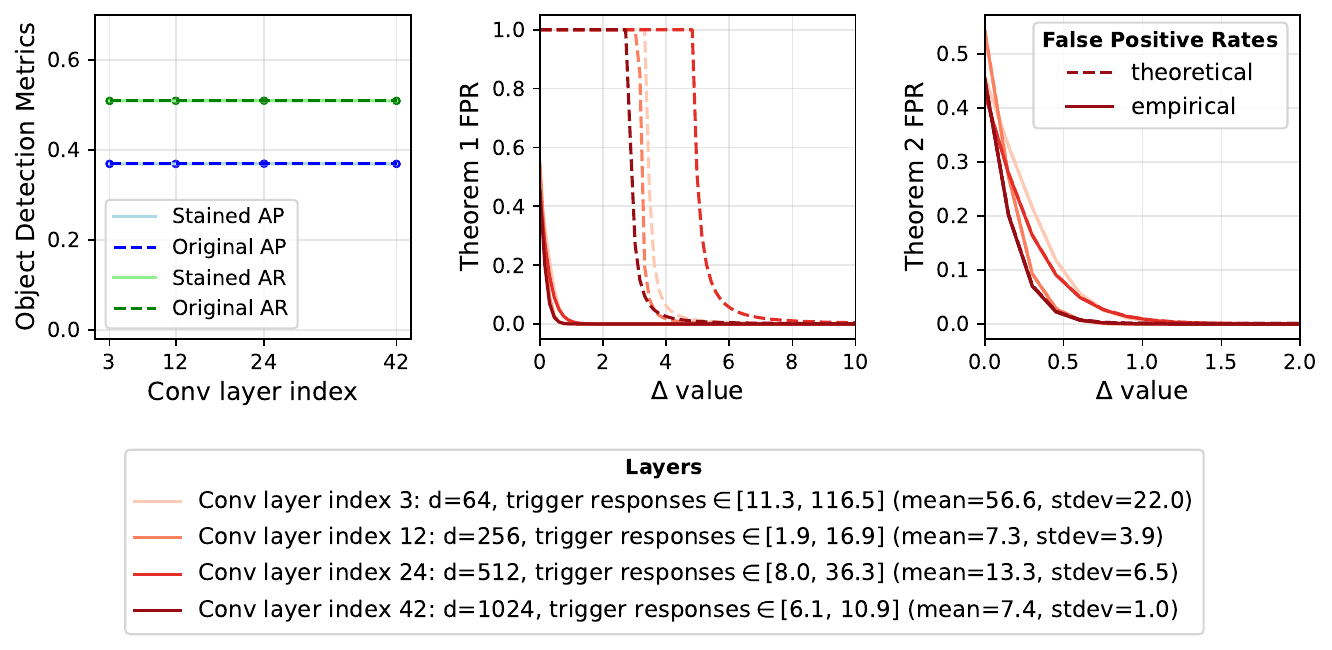}
    \caption{Staining Faster-RCNN}
    \label{fig:stain_internal_fasterrcnn}
\end{figure}

\begin{figure}[h]
    \centering
    \includegraphics[width=1\linewidth]{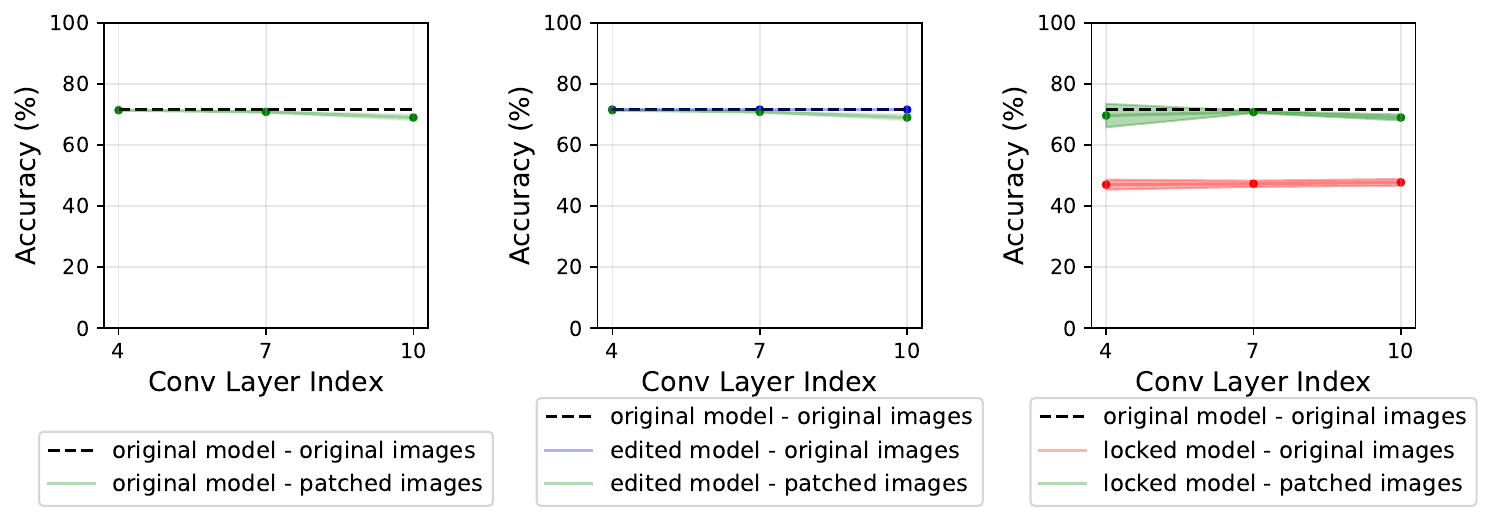}
    \caption{Internal lock for VGG16}
    \label{fig:locking_internal_vgg16}
\end{figure}

\begin{figure}[h]
    \centering
    \includegraphics[width=1\linewidth]{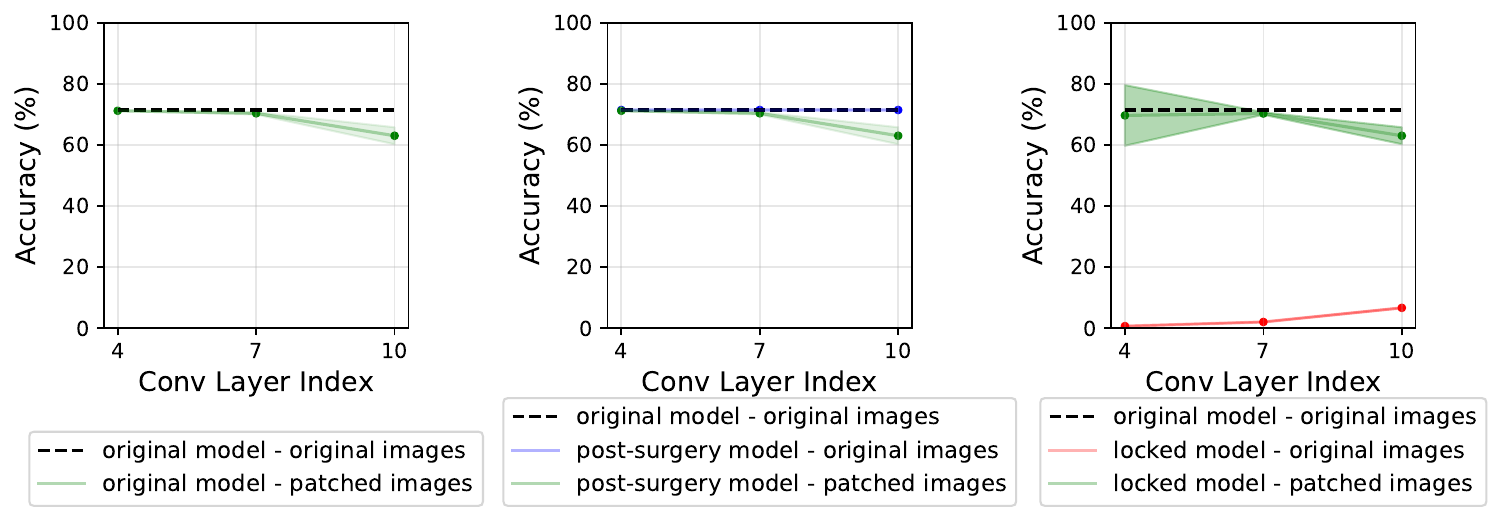}
    \caption{Squeeze-and-excite lock for VGG16}
    \label{fig:locking_sne_vgg16}
\end{figure}

\begin{figure}[h]
    \centering
    \includegraphics[width=1\linewidth]{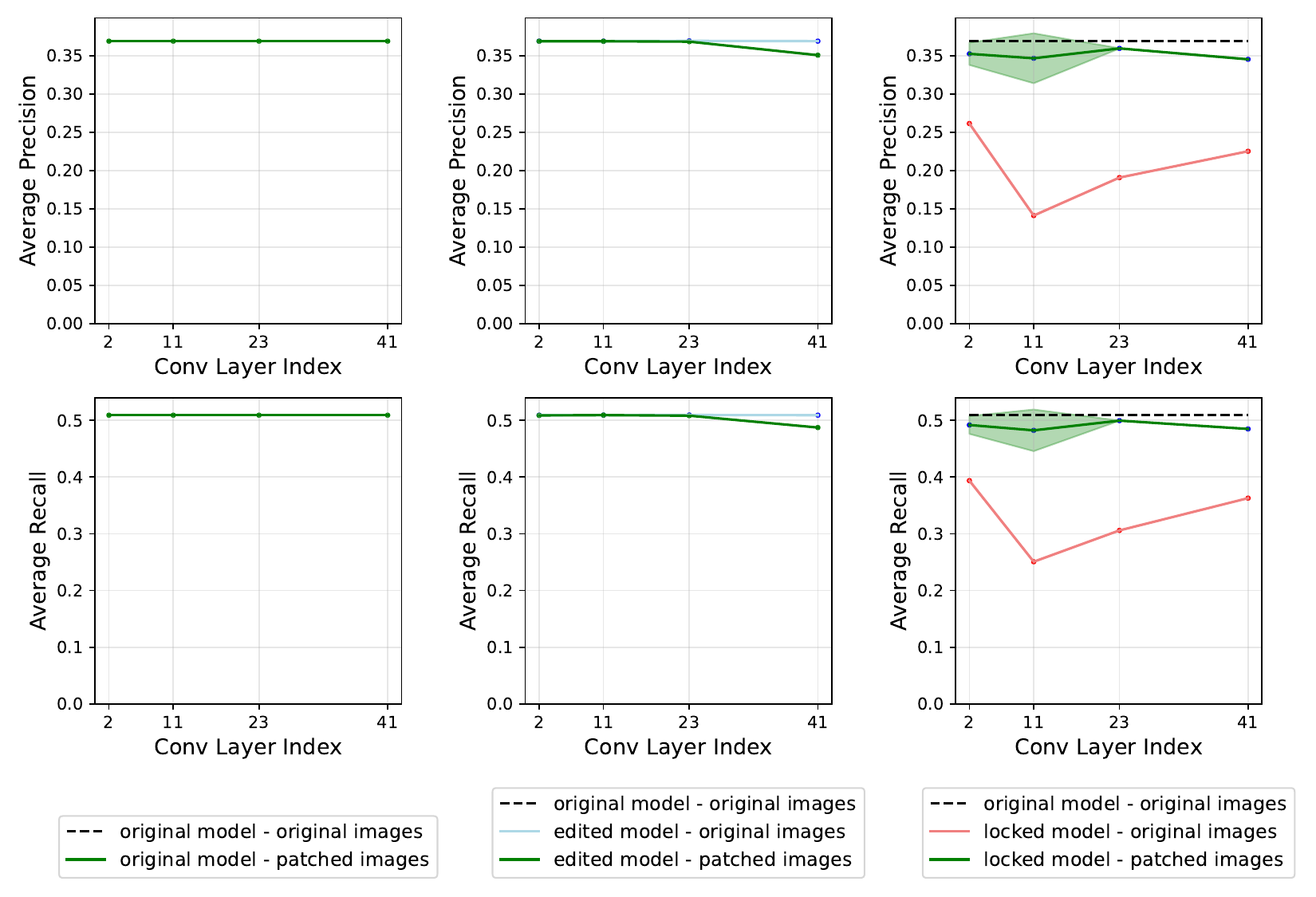}
    \caption{Squeeze-and-excite lock for Faster-RCNN}
    \label{fig:locking_sne_fasterrcnn}
\end{figure}

Here, we include additional figures to support key results in \cref{sec:experiments} of the main text. These include,
\begin{itemize}
    \item \cref{fig:stain_internal_vgg16} for staining VGG-16
    \item \cref{fig:stain_internal_fasterrcnn} for staining Faster-RCNN
    \item \cref{fig:locking_internal_vgg16} for internal lock of VGG-16
    \item \cref{fig:locking_sne_vgg16} for squeeze-and-excite lock of VGG-16
    \item \cref{fig:locking_sne_fasterrcnn} for squeeze-and-excite lock of Faster-RCNN

\end{itemize}

\begin{figure}[h]
    \centering
    \includegraphics[width=1\linewidth]{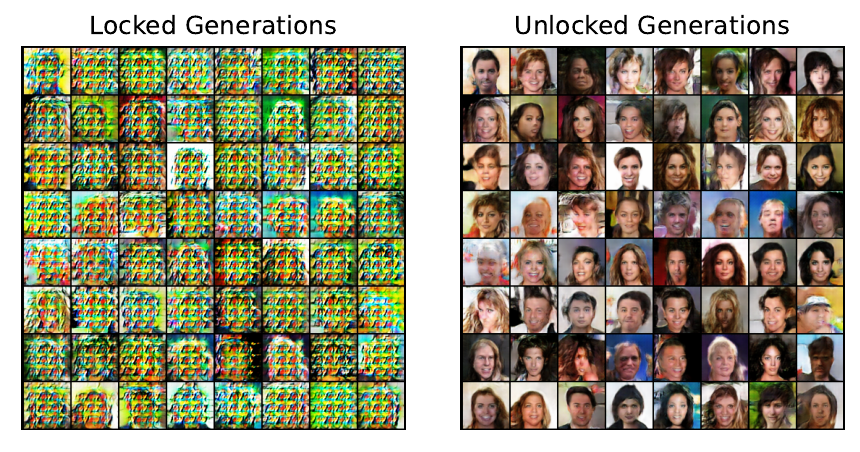}
    \caption{Locked and unlocked generations of DC-GAN}
    \label{fig:lock_images_dcgan}
\end{figure}

\section{Extension to other computer vision models}\label{sec:lock-extensions}

The staining and locking mechanisms introduced in this paper can be extended to various kinds of computer vision models, including but not limited to (1) generative models, e.g. GANs, and (2) vision transformer architectures. Here, we provide some examples. 

\subsection{Staining and Locking DC-GAN}\label{sec:gan-algorithms}

For DC-GAN \cite{radford2015unsupervised}, the generator model is the most vital for image generation tasks. The discriminator is a binary classification model that can be locked either via the internal lock (\cref{alg:lock:internal}) or squeeze-and-excite lock (\cref{alg:lock:squeeze-excite}) by adding a Sq-Ex block. To demonstrate that we can extend the fundamental components and principals of our staining and locking mechanism to image generation models, we focus on staining and locking the generator of one of the most fundamental form of GANs, the DC-GAN.

\boldblock{Staining DC-GAN}
For staining, the method of staining convolutional layers (\cref{alg:stain:convolution}) can be easily extended to transposed convolutional layers, where instead of using the whole kernel $v\in\mathbb{R}^{c_j\times\kappa_j\times\kappa_j}$ as the detector, we only use part of the kernel $v'\in\mathbb{R}^{c_j\times1\times1}$. The example shown in \cref{fig:stain_lock_dcgan} is also a further simplified version of \cref{alg:stain:convolution}, where we rely solely on detector response with $\alpha=1$ and $\beta=0$.

\boldblock{Locking DC-GAN}
For locking, we implement \cref{alg:lock:gan}, where we added additional dimensions to the input random noise vector. A trigger patch of values can be placed into these additional dimensions to unlock the model. The lock is achieved by adjusting one kernel of the first transposed convolutional layer and parameter values of one dimension for the subsequent batch norm layer. These adjustments ensure positive activations from the kernel to the following transposed convolutional layer, where weights $p'$ that can be affected are optimised through gradient descent to destroy to generation performance. By adding the trigger patch, the kernel, filled with the sampled detectors, will provide a fixed output (here $\Delta$) that will result in a large negative response from the batch norm and zero response after the subsequent non-linear activation. This process will effectively turn off the $p'$ and restore a level of model performance. In \cref{fig:stain_lock_dcgan}, we provide more examples of locked and unlocked generations, which clearly demonstrates the effectiveness of the locking and unlocking mechanism.

\subsubsection{Experimental setup}\label{sec:gan:experimentsetup}

\boldblock{Staining DC-GAN}
We insert non-additive stains individually into several transposed convolutional layers of the DC-GAN generator using a modified \cref{alg:stain:convolution}, as discussed above. In each case, the transposed convolution kernel with the least $l^1$ norm weight vector was replaced with the detector neuron, and the optimised trigger is in the form of a $d$-dimensional noise vector. We evaluate the theoretical and empirical FPRs of Theorems \ref{thm:geometric} and \ref{thm:datadriven} of the stain in the same manner as in \cref{sec:experiments}, adjusted for transposed convolution instead of standard convolution. To ensure the stain has minimal impact on the generator performance, we evaluate discriminator losses for both stained and original generators.

\boldblock{Locking DC-GAN}
We provide an example of DC-GAN lock by inserting an non-additive stain into the first transposed convolutional layer, which is modified to take in $d+d_a$ dimensional inputs, where $d_a=20$. The locking mechanism is then inserted into the following batch-norm and transposed convolutional layers according to \cref{alg:lock:gan} with parameters $\Delta=30, \xi=10$. Since the staining layer is different to those evaluated in the prior section, we also evaluate empirical FPRs of Theorems \ref{thm:geometric} and \ref{thm:datadriven} of the stain at layer 0. We evaluated the lock via discriminator losses on a fixed set of randomly sampled noise vectors of dimension $d+d_a$ with or without the $d_a$-dimensional trigger patch inserted. To further validate the effectiveness of the lock, we generate images with both the locked and unlocked models for visual confirmation.

\begin{figure}[h]
    \centering
    \includegraphics[width=0.8\linewidth]{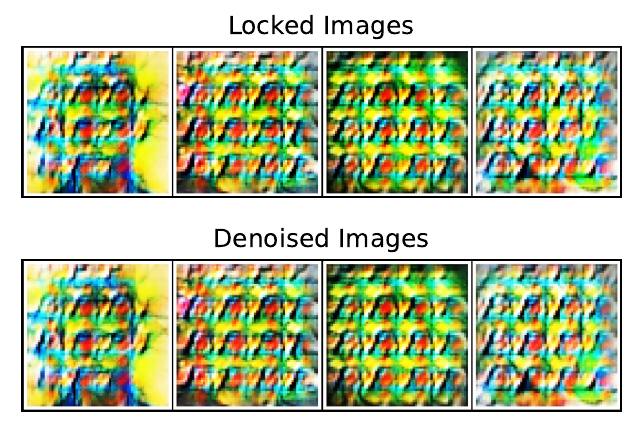}
    \caption{Example of denoising locked images with NAFNet}
    \label{fig:lock_denoise_dcgan}
\end{figure}

\boldblock{Verification of lock for DC-GAN}
For image generation, we can also verify the lock via attempting to recover good or original generations via denoising or image restoration methods. We provide an example attempt at denoising with NAFNet \cite{chu2022nafssr} in \cref{fig:lock_denoise_dcgan}.

\begin{algorithm}
    \caption{DC-GAN lock}\label{alg:lock:gan}
    \small
    \SetKwInOut{Input}{Input}\SetKwInOut{Output}{Output}
    \Input{Trained generator $\mathcal{G}$ and discriminator $\mathcal{D}$
        \\
        Parameters $(k, \Delta)$ as in Alg.~\ref{alg:stain:perceptron}
        \\
        Scaling parameter $\xi$
        \\
        Number of additional noise dimensions $d_a$
    }

    Let $W_0\in\mathbb{R}^{c_0\times d \times \kappa \times \kappa}$ be the weight of the initial transposed convolutional layer, with kernel shape of $\kappa \times \kappa$ and stride 1. The input into this layer is a random noise vector of dimension $d$.

    Sample a detector vector $\pi \in \mathbb{R}^{d_a}$ from $\mathcal{U}(\mathbb{S}^{d_a - 1})$, and use it to construct a kernel $v \in \mathbb{R}^{(d+d_a)\times \kappa \times \kappa}$, where from $(v)_{1,...,d}=0$ and $(v)_{d+1,...,d+d_a}\in\mathbb{R}^{d_a\times\kappa\times\kappa}$ is made of $\kappa\times\kappa$ duplicates of $\Delta \pi$, where $\Delta$ is a constant. 
    
    Here, the detector vector $\pi$ is also the trigger patch to be added to the additional dimensions $d_a$ of the random noise input to unlock the model.
    
    Replace the initial transposed convolutional layer with zero weight $W_0'\in\mathbb{R}^{c\times (d+d_a) \times \kappa \times \kappa}$ where we added $d_a$ dimensions to the noise vector inputs. 
    Replace weights corresponding to the first $d$-dimensions of $W_0'$ with weights from $W_0$. Replace kernel with index $k$ (one out of a total of $c_0$ kernels) with $v$.

    Let $\mu_0, \sigma_0^2, w_0, b_0 \in \mathbb{R}^{c}$ be the mean, variance, weight and bias of the layer $0$ batch normalisation after the initial transposed convolutional layer.
    \begin{minipage}{0.9\linewidth}
        \begin{itemize}
            \item replaced $\mu_0$ with $s\mu_0' = \mu_0 + \xi$, where $\xi$ is a positive offset to prevent negative detector responses
            \item replace $w_0$ with $w_0'$ where $(w_0')_k = \frac{(s - (b_0)_k)(\sigma_0)_k}{\Delta - (\mu_0)_k}$, $s$ is a large negative constant
        \end{itemize}
    \end{minipage}

    Let $W_1\in\mathbb{R}^{c_1\times c_0 \times \kappa \times \kappa}$ be the weight of the second transposed convolutional layer, where a portion of the weights $p\in\mathbb{R}^{c_1\times\kappa\times\kappa}$ is affected by dimension $k$ in the prior layer.

    Optimise a tensor $p'$ such that when the corresponding portion of weights in $W_1$ is replaced by $p'$, the discriminator outputs of $\mathcal{D}(\mathcal{G}'(x))$ will minimise the binary cross-entropy loss to all false predictions.

    Replace the portion of weights corresponding to $p$ in $W_1$ with $p'$
    
    \Output{~Locked generator $\mathcal{G}$ and trigger patch $\pi$}
    
\end{algorithm}

\subsection{Staining and Locking ViT}\label{sec:vit-algorithms}

To demonstrate that our staining and locking mechanisms can extend beyond the simple convolutional architectures, we present examples on staining and locking vision transformer architectures like ViT-16-B \cite{dosovitskiy2020image}.

\boldblock{Staining ViT}
ViT encoder blocks contain multi-layer perceptron (MLP) modules.
Any neuron in an MLP layer can be stained using \cref{alg:stain:perceptron}.
The only modification to this algorithm required to extend it to ViTs is that we also have to choose a specific token to optimise and read the detector responses.

\begin{figure}[h]
    \centering
    \includegraphics[width=\linewidth]{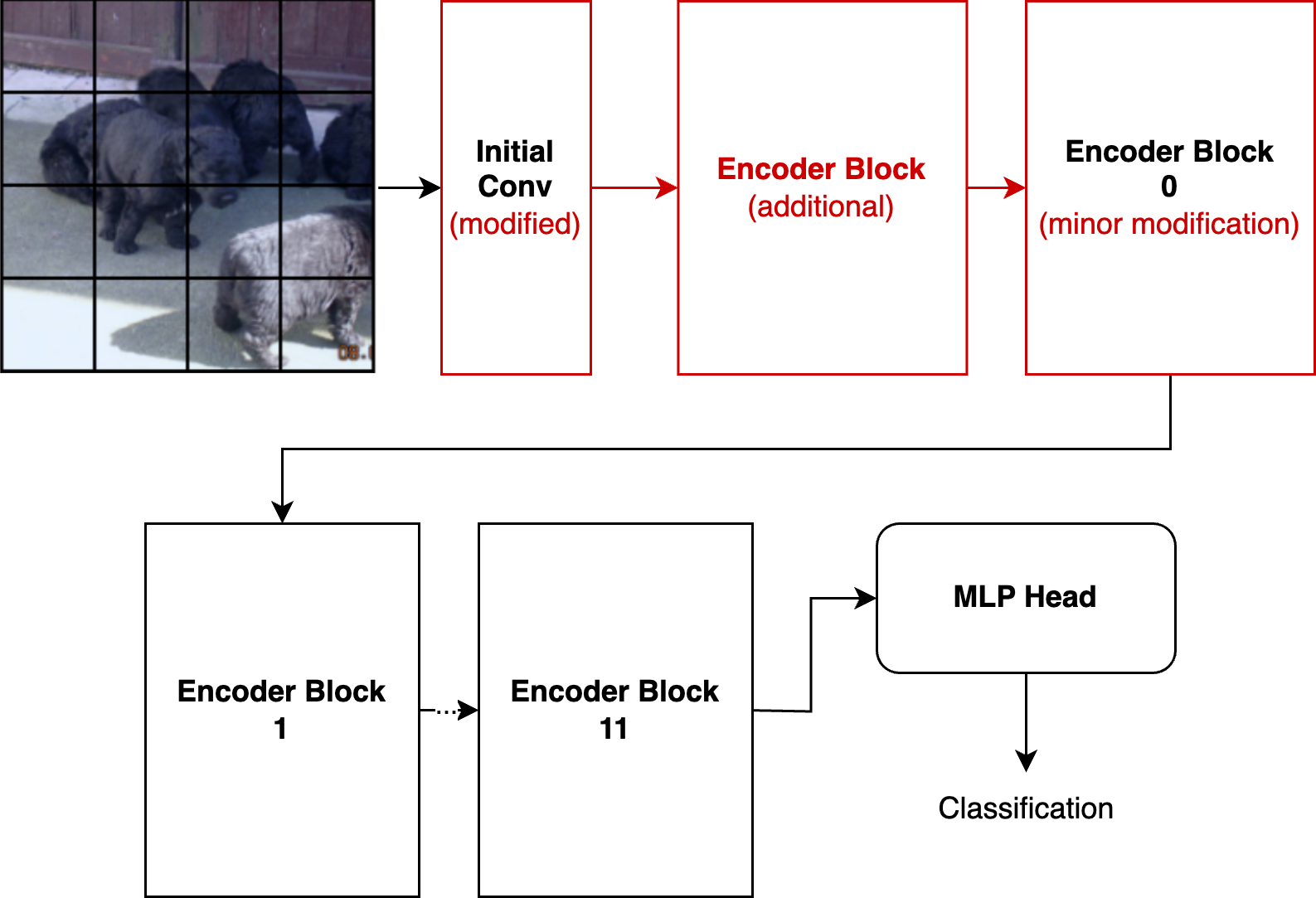}
    \caption{Basic illustration of example ViT lock}
    \label{fig:schema_macro_vit}
\end{figure}

\begin{figure}
    \centering
    \includegraphics[width=0.85\linewidth]{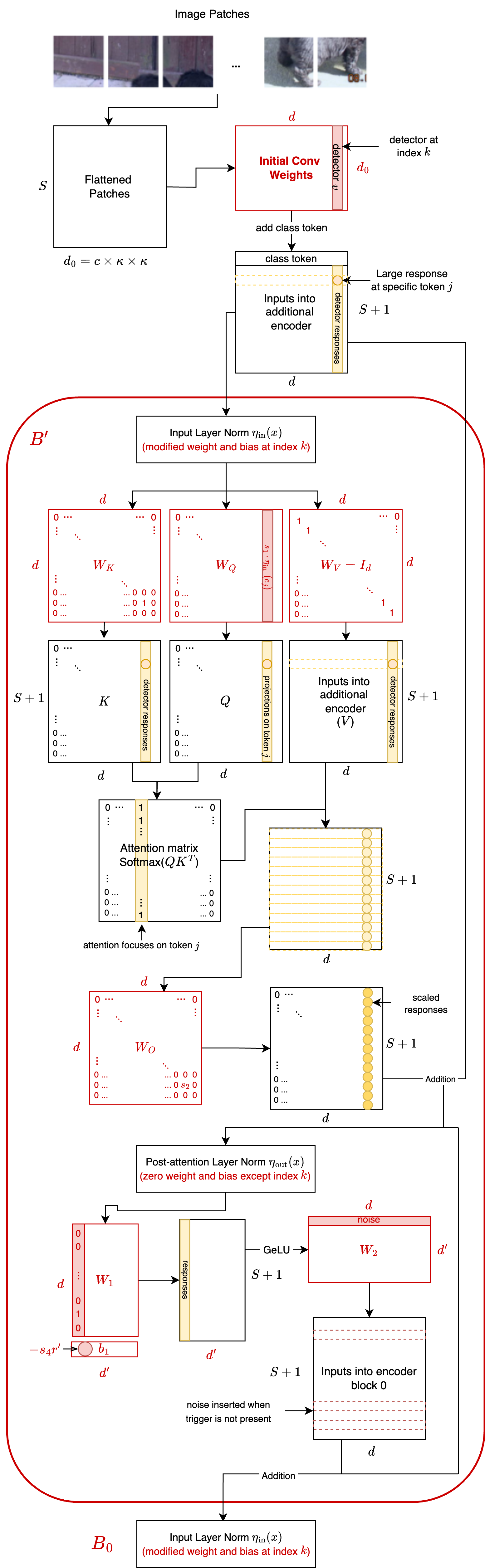}
    \caption{Detailed illustration of example ViT lock}
    \label{fig:schema_full_vit}
\end{figure}

\boldblock{Locking ViT}
Since ViT contains an initial convolutional layer to transform image patches into tokens.
We can stain a single kernel of the layer  (\cref{alg:stain:convolution} steps 1-4). We identify this kernel as the $k$-th kernel. Then, we can add an additional encoder block after the convolutional layer and before the first original encoder block. This additional block will contain two layer-norms, an attention module and an MLP module. See \cref{alg:lock:vit} and \cref{fig:schema_full_vit} for details.
The weights of the attention module are designed such that it only focuses on the inputs of a single token and only outputs non-zero responses for the $k$-th dimension. 
A detector neuron is implanted into the first layer of the MLP module to respond to the output from the stained kernel in the convolutional layer.
When activated, the signal from this detector acts as a switch to a noise vector implanted in the second layer of the MLP module.
The noise is optimised via gradient descent to reduce the model's classification performance significantly. A basic illustration of the lock can be found in \cref{fig:schema_macro_vit}

When the trigger is not present, the response from the detector neuron will be positive for a portion of the tokens, introducing noise to the MLP outputs and subsequent residual stream. When the trigger is present, the response from the detector neuron in the first layer of the MLP will be large and negative, which will be zero after the GeLU activation, preventing the noise from the disruptor in the second MLP layer from being introduced into the model.
The input layer norm in the following encoder block is also slightly modified to reduce the propagation of unnecessary noise when the model is unlocked.
There also exists a balance between the functionality of the lock and preserving the input to the original encoder blocks, due to the existence of multiple layer norms that are susceptible to distortion with large noises.
In \cref{alg:lock:vit}, this balance is managed by the various scaling factors.

\subsubsection{Experimental setup}\label{sec:vit:experimentsetup}

\boldblock{Staining ViT}
We insert non-additive stains individually into several MLP modules of the ViT-16-B model with a modified \cref{alg:stain:perceptron}. In each case, the layer 1 neuron with the least $l^1$ norm weight vector was replaced with the weight vector. The trigger optimisation is limited to a specific token. In \cref{fig:stain_lock_vit}, we choose token 1 (the first token that is not the classification token), which corresponds to patch $(0,0)$ in the input image.  We evaluate the theoretical and empirical FPRs of Theorems \ref{thm:geometric} and \ref{thm:datadriven} of the stain in the same manner as in \cref{sec:experiments}, adjusted for MLP inputs, where we look at inputs to all tokens. The impact of the stain on model performance is also evaluated by evaluating the accuracy of the stained and original models.

\boldblock{Locking ViT}
We provide an example of ViT lock by inserting a non-additive stain into the initial convolutional layer (\cref{alg:stain:convolution} steps 1-4) that has the least $l^1$ norm weight vector, with trigger optimised on patch $(0,0)$. Then we construct the locking mechanism as detailed in \cref{alg:lock:vit} with parameters $\alpha=0.025$, $s_0=1$, $s_1=100$, $s_2=0.1$, $s_3=100$, $s_4=0.9$. We evaluate the empirical FPRs of Theorems \ref{thm:geometric} and \ref{thm:datadriven} of the stain on the initial convolutional layer. The function and impact of the lock is evaluated by measuring the accuracy of the original, locked and unlocked models.

\begin{algorithm}
    \caption{Vision Transformer (ViT) lock}\label{alg:lock:vit}
    \small
    \SetKwInOut{Input}{Input}\SetKwInOut{Output}{Output}
    
    \Input{Trained vision transformer $\mathcal{V}$
            \\
            Parameters $k$ as in Alg.~\ref{alg:stain:perceptron}
            \\
            Scaling parameters $\alpha, s_0, s_1, s_2, s_3, s_4$
            \\
            Image coordinates (a,b) for the trigger patch
            \\
            Example images with ground truth labels
        }

    Let $W_0 \in \mathbb{R}^{c \times d \times \kappa \times \kappa}$ be the weight of the initial convolutional layer, with kernel shape $\kappa \times \kappa$.

    Sample a kernel $v \in \mathbb{R}^{c \times \kappa \times \kappa}$ from $\mathcal{U}(\mathbb{S}^{c \kappa^2 - 1})$, viewed as a tensor with shape $(c, \kappa, \kappa)$.

    Optimise the trigger input $x^* \in \arg\max_{z \in S} r_{(a, b)}(v * \phi(z))$.
    
    Replace $(W_0)_{k, \cdot, \cdot, \cdot}$ with $\alpha v$, and entry $k$ of $b$ with $0$, where $\alpha$ is a constant scaling factor.

    Define the trigger patch $\pi$ as the part of the trigger input $x^*$ which is in the receptive field of $r_{(a, b)}(v * \phi(x^*))$. This patch corresponds to trigger token $m$

    Optimize a noise vector $\nu$ through gradient descent such that when added to the output of the first convolutional layer, the ViT output $V(x)$ will generate the maximum cross-entropy loss with ground truth labels of the set of example images. 

    Construct an additional encoder block $B'$ consisting of
    \begin{minipage}{0.9\linewidth}
    \begin{itemize}
        \item an input layer norm $\eta_{\text{in}}(x):\mathbb{R}^d\rightarrow \mathbb{R}^d$ with $w_{l,\text{in}}', b_{l,\text{in}}'\in \mathbb{R}^d$ which are duplicates of $w_{l,\text{in}}, b_{l,\text{in}}\in \mathbb{R}^d$ of the first encoder block input norm with index $k$ weight replaced with 1 and bias replaced with a constant $s_0$

        \item an attention module with key, value and query projection matrices $W_K, W_V, W_Q\in\mathbb{R}^{d\times d}$, and output projection matrix $W_O\in\mathbb{R}^{d\times d}$,  where 
            \begin{itemize}
                \item column $k$ of $W_K$ is a vector $s_1\cdot\eta_{\text{in}}(e_j)$, where $s_1$ is a large constant and $e_j$ is the $j$-th row of the positional embedding $e$ corresponding to the $j$-th token. All other values of $W_K$ are zero.
                \item $(W_Q)_{k,k} = 1$, while all other values of $W_Q$ are zero.
                \item $W_V$ is an identity matrix of dimension $d$
                \item $(W_O)_{k,k} = s_2$ where $s_2$ is small constant., while all other values of $W_O$ are zero.
            \end{itemize}

        \item a post-attention layer-norm $\eta_{\text{out}}:\mathbb{R}^d\rightarrow \mathbb{R}^d$ $w_{l,\text{out}}', b_{l,\text{out}}'\in \mathbb{R}^d$ where all values apart from $(w_{l,\text{out}}')_k=-s_3$ where $s_3$ is a constant.

        \item a multi-layered perception module consisting of two linear layers with weights $W_1\in \mathbb{R}^{d\times d'}, W_2\in\mathbb{R}^{d'\times d}$ and biases $b_1\in\mathbb{R}^{d'}, b_2\in\mathbb{R}^d$, where
            \begin{itemize}
                \item $(W_1)_{1,k}=1$ and all other values of $W_1$ is zero.
                \item $(b_1)_k = - s_4 r'$ where $r'$ is the maximum response of column $k$ of $W_1$ to trigger patched inputs
                \item replace row $k$ of $W_2$ with noise $\nu$, all other values of $W_2$ and $b_2$ are zero
            \end{itemize}

    \end{itemize}
    \end{minipage}

    Insert $B'$ between the initial convolutional layer and the first encoder block $B_1$.

    Replace index $k$ values of $B_1$'s input layer norm weight and biases $w_{l,\text{in}}, b_{l,\text{in}}$ with the value of 0.

    \Output{~Locked model $\mathcal{V}$ and trigger patch $\pi$}
    
\end{algorithm}

\subsection{Lock position and obfuscation for GAN and ViT}
For the examples provided to locking GAN and ViT, the example locking mechanism conceals the lock within similar structures to that of the original model: (1) for locking GAN, we only changed the input dimension of the random noise vector, which structurally is the first transposed convolutional layer, (2) for locking ViT, the locking mechanism is concealed within an additional encoder block that is structurally identical to any other encoder block. 
Numerous approaches could be used to obfuscate and hide these components of the locks, although for brevity we do not explore this here.

For simplicity, we present the algorithms for locking ViT and GANs with the lock in a single position within the network.
It is a direct generalisation of the algorithms to place the locks in different parts of the model, although we do not pursue the details of this here.

\subsection{Staining Swin-Transformers and Diffusion models}

The architecture of Swin Transformers \cite{liu2021swin} is closely related to that of ViTs, and diffusion models like DDPM are built around CNNs \cite{ho2020denoising}. Therefore, the same methods of staining for ViT in Section~\ref{sec:vit-algorithms} and CNN in Algorithm~\ref{alg:stain:convolution} can be applied respectively.

\subsubsection{Experimental Setup}

\begin{figure}[h]
    \centering
    \includegraphics[width=\linewidth]{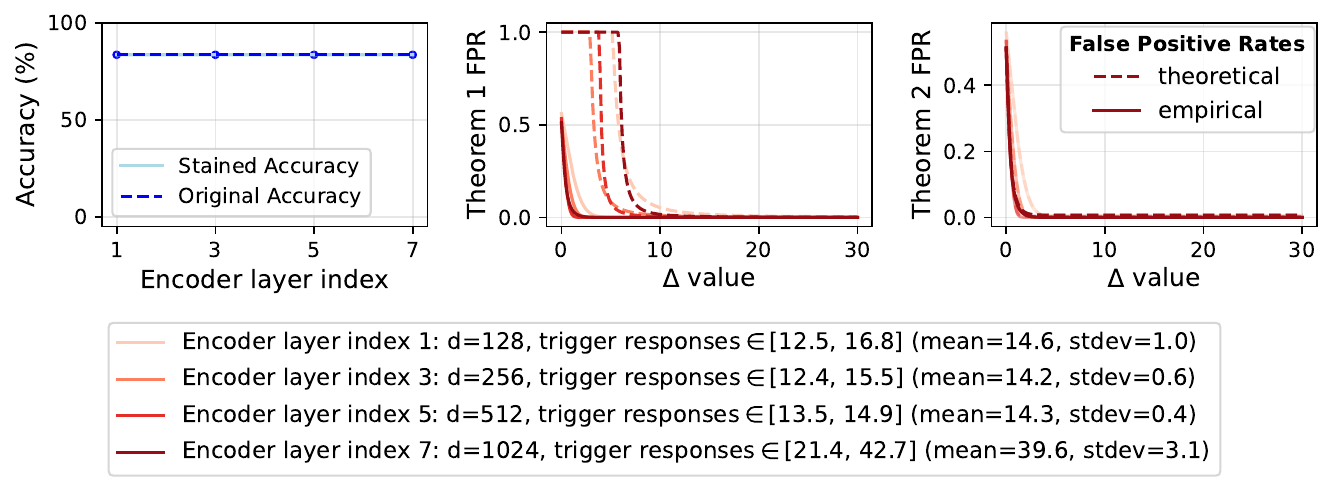}
    \caption{Staining Swin-b}
    \label{fig:stain-swin}
\end{figure}

\boldblock{Staining Swin-b} We insert non-additive stains individually into several MLP modules of the Swin-b encoder with the same method as Section~\ref{sec:vit-algorithms}. We evaluate the theoretical and empirical FPRs of Theorems \ref{thm:geometric} and \ref{thm:datadriven} of the stain and the impact of the stain on accuracy in the same way as Section~\ref{sec:vit-algorithms}. The results are shown in Figure~\ref{fig:stain-swin}. The mean accuracy difference across 40 stains and 4 layers is $-0.55\%$ ($0\%, -1\%, -1.4\%, 0.2\%$ across stains for each of layers 1-7, respectively).

\begin{figure}[h]
    \centering
    \includegraphics[width=\linewidth]{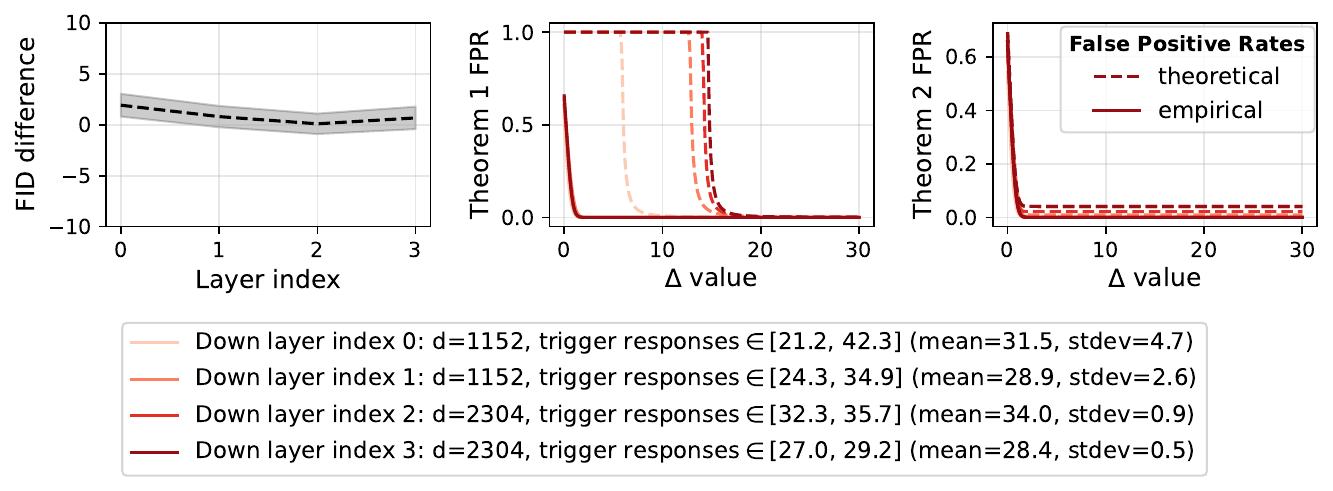}
    \caption{Staining DDPM}
    \label{fig:stain-ddpm}
\end{figure}

\boldblock{Staining DDPM} We insert non-additive stains individually into the convolutional layer of several downwards residual blocks of the U-Net backbone of DDPM. We evaluate the theoretical and empirical FPRs of Theorem \ref{thm:geometric} and \ref{thm:datadriven} of the stain in the same way as for convolutional network staining in Section~\ref{sec:experiments}. Since DDPM performs an image generation task, we measure the Fréchet inception distance (FID) score between generated and real images. The results are shown in Figure~\ref{fig:stain-ddpm}. We measure FID for the original and stained models and compare the differences ($1.3\%$ difference across 40 stains and 4 layers; $3\%, 1.2\%, 0.14\%, 1.0\%$ across 40 stains for layers 0-3, respectively).

\section{Extended discussion on security of stains and locks}
\label{sec:supplementary:robustness-discussion}

\begin{figure}[h]
    \centering
    \includegraphics[width=\linewidth]{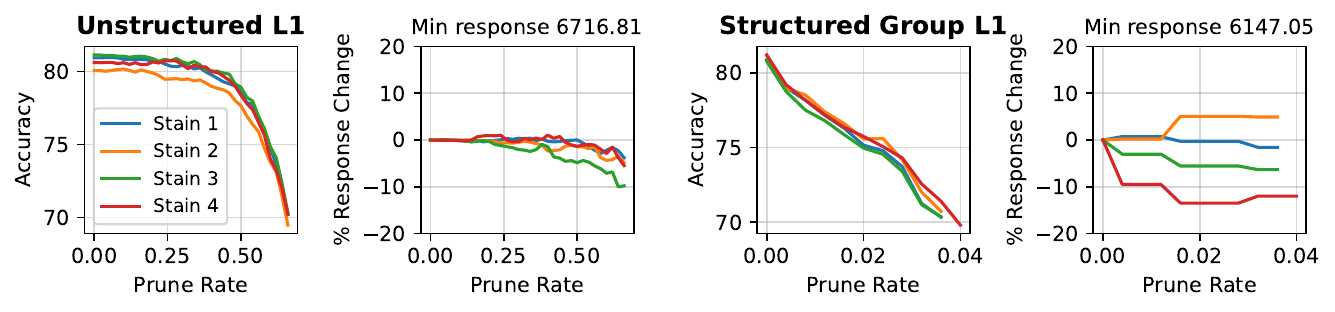}
    \caption{Survival of stains/locks under pruning}
    \label{fig:robustness:pruning}
\end{figure}

\boldblock{Pruning attacks} 
If the stain/detector is applied additively to one or several key neurons in the model (e.g. on the `golden lottery ticket'), then they cannot be pruned without severely impairing the model. If the lock disruptor is pruned,  then the model is simply permanently locked and therefore has its performance impaired. Additional experiments shown in Figure~\ref{fig:robustness:pruning} demonstrate that stains/locks survive pruning, and structured and unstructured standard $L_1$ pruning have relatively small impact on the performance of a detector ($\%$ change in neuron's response to triggers). 

\boldblock{Fine-tuning and weight perturbation attacks} 

The robustness of our stains to weight perturbations such as fine-tuning follows as a direct consequence of Theorem~\ref{thm:geometric}, which may be expressed as the following result.

\begin{corollary}[Robustness to weight perturbations] Let the map $\phi: S\rightarrow\mathbb{R}^d$ and the vector $w\in\mathbb{R}^d$ be defined as in Theorem \ref{thm:geometric}. Suppose that test data $x$ are independently sampled from a distribution $\mathcal{D}$ on $S$ such that $\mathbb{E}_x[\phi(x)] = \mu$ and $\operatorname{Cov}(\phi(x))$ has eigenvalues $\{\lambda_i\}_{i=1}^{d}$. Pick any $\Delta > \|\mu\|$ and $\delta   \in(0, \Delta - \|\mu\|)$.
Let $\hat{w}\in\mathbb{R}^d$, $\hat{\phi}: S\rightarrow\mathbb{R}^d$ be such that the following hold true with probability one
    \begin{equation}\label{eq:corr:finetuningassumption}
        |\hat{w} \cdot \hat{\phi}(x)-w \cdot \phi(x)| \leq \delta.
    \end{equation} 

Then
    \[
    \begin{aligned}
    P(x\sim\mathcal{D} & : \ \hat{w}\cdot \hat{\phi}(x)>\Delta )\\
    & \leq
        \frac{\sum_{i=1}^{d} \lambda_i}{2(\Delta - \|\mu\| - \delta)^2} \frac{d-1}{d+1} \Big(\frac{\Gamma(\frac{d}{2})}{\Gamma(\frac{d+1}{2})} \Big)^2.
    \end{aligned}
    \]
\end{corollary}

Indeed, consider events
\[
E_1: \ \hat{w} \cdot \hat{\phi}(x) \leq \Delta, \ E_2: \ {w} \cdot \phi(x) \leq \Delta-\delta,
\]
and
\[
E_3: \ |\hat{w} \cdot \hat{\phi}(x)-w \cdot \phi(x)| \leq \delta.
\]
If events $E_2$ and $E_3$ occur, then event $E_1$ occurs too. Hence
\[
\begin{aligned}
& P(E_1) \geq P(E_2 \ \& \ E_3) \geq 1 - P(\mathrm{not} \ E_2) - P(\mathrm{not} \ E_3) \\
& = 1 - P(\mathrm{not} \ E_2).
\end{aligned}
\]
The second inequality in the expression above follows from the classical De Morgan's law, and the last equality is due to the fact that event $E_3$ holds true with probability one. Therefore
\[
P(\mathrm{not} \ E_1) \leq P(\mathrm{not} \ E_2),
\]
and the probability $P(\mathrm{not} \ E_2)$,  according to Theorem \ref{thm:geometric}, is bounded above by
\[
\frac{\sum_{i=1}^{d} \lambda_i}{2(\Delta - \|\mu\| - \delta)^2} \frac{d-1}{d+1} \Big(\frac{\Gamma(\frac{d}{2})}{\Gamma(\frac{d+1}{2})} \Big)^2.
\]

\begin{figure}[h]
    \centering
    \includegraphics[width=\linewidth]{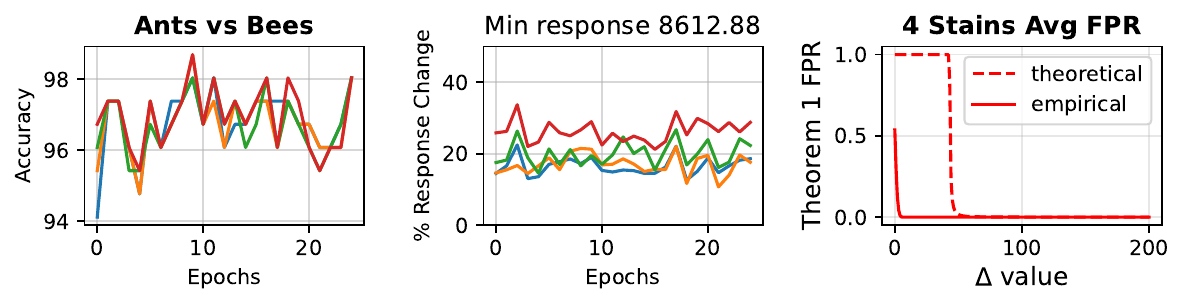}
    \caption{Fine-tuning}
    \label{fig:robustness:fine-tuning}
\end{figure}

In fine-tuning, assumption ~\eqref{eq:corr:finetuningassumption} typically holds with $\delta$ small relative to $\Delta$ (which represents the maximum projection possible), ensuring that the model's performance does not degrade.
Consequently, the behaviour of the stain and lock after fine-tuning is expected to persist.
The  experiments in Figure~\ref{fig:robustness:fine-tuning} illustrate empirical justifications stemming from the theory.

\begin{figure}[h]
    \centering
    \includegraphics[width=\linewidth]{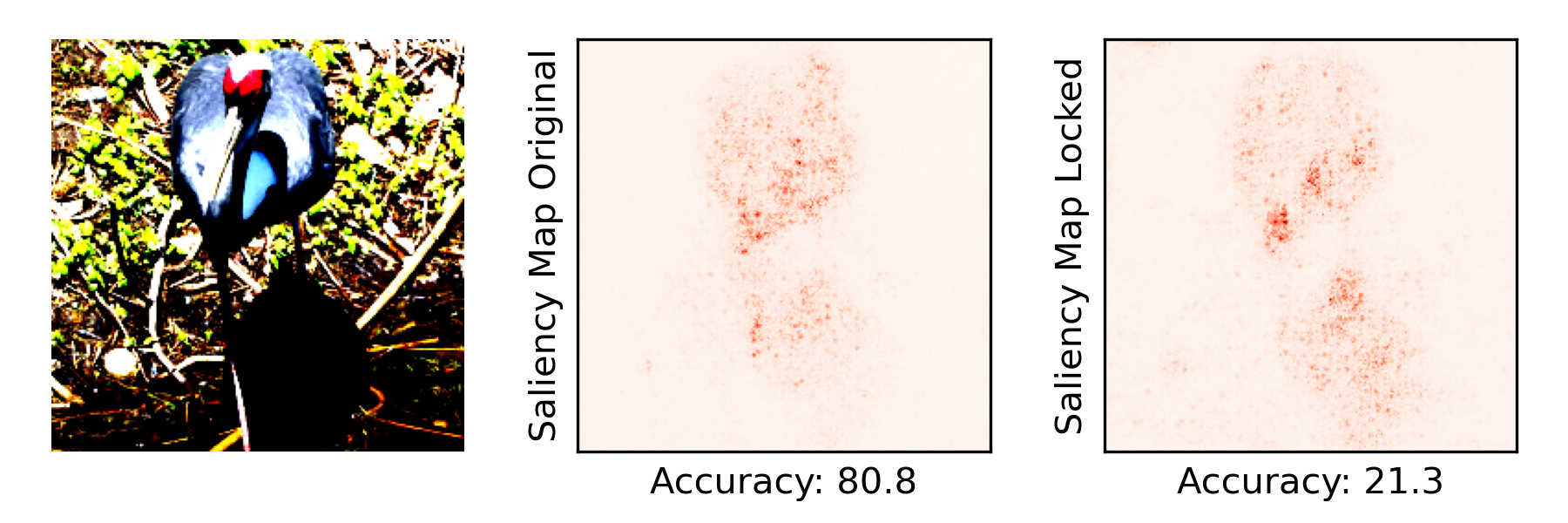}
    \caption{Saliency maps}
    \label{fig:robustness:adversarial-image-analysis}
\end{figure}

\boldblock{Adversarial image analysis}
When the trigger patch is present, the model is unlocked. When the model is locked, no trigger patches are present. Hence the detector neuron behaves like any other neuron, so there is no sign in the model's activations that such a patch is required (or where or how large it should be, or what values it should take).
Saliency analysis of perturbations does not apply to images without the trigger in them. We demonstrate this for a classification CNN in Figure~\ref{fig:robustness:adversarial-image-analysis} -- since the detector is convolutional, the trigger can be placed anywhere in the image, but the saliency map does not show this.

\boldblock{Leaked keys}
As with any scheme involving a key, security cannot be guaranteed if the key is leaked.  Low collision probability (below) enables detection of abnormal usage patterns (like what most banks do) which may correlate with the key leakage. These risks, nevertheless, should be considered. Our method uniquely enables replacing the compromised stain/lock post-deployment to multiple clients on the fly without retraining.

\boldblock{Forged keys}
Our work reveals that forging attacks are trivially feasible on \emph{all known staining methods}. This is a key contribution of our work which it is important for the community to be aware of. Despite this, inserting the forged stain requires access to the model's weights.

\boldblock{Obfuscation}
Even in the `vanilla' setting presented here, the lock provides an asymmetrically difficult task for the thief while requiring little work from the owner. We agree that obfuscation is required for practical implementations, and will comment further on this in the discussion. 

\boldblock{Multi-client distribution and collisions}
Since the detector neuron weights are sampled uniformly from the sphere, the probability of sampling two with dot product greater than $\theta$ is less than $\exp(-\frac{d \theta^2}{2})$ in a feature space of dimension $d$~\cite{ball1997elementary}. Hence, the number of clients who can be handled grows exponentially with the feature space dimension.

\end{document}